\documentclass[10pt]{article}

\usepackage[authoryear,square]{natbib}

\usepackage{graphicx}
\usepackage{subcaption}

\usepackage{booktabs}
\usepackage{multirow}
\usepackage{longtable}

\usepackage{amsmath}
\usepackage{bm}
\usepackage{amsthm}
\usepackage{amsfonts}
\usepackage{amssymb}
\usepackage{mathtools}
\mathtoolsset{showonlyrefs=false}
\newtheorem{theorem}{Theorem}[section]

\usepackage[
    vlined, 
    ruled, 
    linesnumbered 
]{algorithm2e}

\usepackage{xcolor}

\usepackage[hidelinks]{hyperref}

\usepackage{takeuchi_mcr}

\title{\vspace{0mm}Selective Inference for Deep Clustering in Latent Spaces}

\date{\today}

\makeatletter
\def\@fnsymbol#1{\ensuremath{\ifcase#1\or
{1}\or 
{\ast}\or 
{2}\or 
{\dagger}\or 
\else\@ctrerr\fi}}
\makeatother

\author{
Eina Mizui\thanks{Nagoya University} ,
Tomohiro Shiraishi\footnotemark[1] \thanks{RIKEN} ,\\
Shunichi Nishino\footnotemark[1] \footnotemark[2] ,
Ichiro Takeuchi\footnotemark[1] \footnotemark[2] \thanks{Corresponding author. e-mail: takeuchi.ichiro.n6@f.mail.nagoya-u.ac.jp}
}

\begin{document}

\maketitle
\thispagestyle{empty}

\begin{abstract}
    \noindent 
    Deep clustering is a powerful approach for discovering meaningful structures in high-dimensional data by learning a low-dimensional latent representation prior to clustering.
Despite its empirical success, assessing the statistical reliability of the resulting clusters remains challenging.
Testing discovered clusters on the same data induces selection bias and invalidates classical $p$-values.
Selective inference (SI) provides a principled framework for correcting this bias, but existing methods focus on clustering performed directly on the observed features.
In this work, we develop an SI framework for deep clustering with a fixed pretrained encoder.
The key challenge is that cluster assignments are determined through a nonlinear transformation from the original data space to the latent space, resulting in a substantially more complex selection process than in conventional clustering.
Our method provides a computationally tractable way to account for this process and enables valid statistical testing of differences between clusters identified in the latent space.
Synthetic experiments demonstrate that the proposed method controls the Type I error rate while achieving higher power than valid but conservative baselines, and genomic applications show that it can identify significant cluster differences while appropriately accounting for selection bias.
Our framework provides a principled approach to quantifying the statistical reliability of structures discovered by deep clustering.

\end{abstract}

\newpage
\section{Introduction}
\label{sec:introduction}

The rapid growth of measurement technologies and computational power has led to the widespread availability of large-scale, high-dimensional datasets across scientific disciplines.
A central challenge in analyzing such data is to uncover latent structures, such as meaningful subgroups, that are often obscured by noise and high dimensionality.
Clustering is a fundamental tool for discovering such latent structures and has become an essential component of data analysis in many scientific domains.

Directly applying distance-based clustering methods such as $k$-means \citep{lloyd1982least} to high-dimensional data can, however, be challenging due to the curse of dimensionality; in particular, distance concentration can reduce the discriminative value of Euclidean distances \citep{beyer1999nearest}.
For this reason, it is common to first construct a lower-dimensional representation of the data and then perform clustering in the resulting representation space.
Classical approaches rely on linear dimensionality-reduction methods such as principal component analysis, whereas more recent approaches employ nonlinear representation learning based on deep neural networks, including autoencoders \citep{hinton2006reducing}.
The resulting family of approaches is commonly referred to as deep clustering \citep{min2018survey,zhou2025comprehensive}.
In this study, we focus on a sequential deep clustering setting, in which a pretrained encoder maps high-dimensional observations into a low-dimensional latent space that is subsequently partitioned by a clustering algorithm such as $k$-means.

Our motivating application is single-cell RNA sequencing (scRNA-seq), where gene expression profiles spanning tens of thousands of dimensions are analyzed to identify biologically meaningful cell populations \citep{zheng2017pbmc}.
Dimensionality reduction followed by clustering is a standard workflow in single-cell analysis \citep{luecken2019current}, and deep representation learning has been used to extract low-dimensional structures that facilitate the identification of cell populations \citep{geddes2019autoencoder}.
Once such data-driven subgroups have been identified, an important downstream task is to determine which genes differ in expression between the discovered populations.
More generally, after deep clustering, one would like to assess whether the identified clusters exhibit statistically supported differences in individual features.
Reliable inference for such differences is particularly important in scientific applications, because apparent cluster-specific signals caused merely by sampling variation may lead to misleading biological interpretations and unnecessary follow-up experiments.

Statistical inference after clustering is challenging because the clusters themselves are selected from the same data used for subsequent testing.
For example, if a standard hypothesis test is applied to compare feature means between two data-driven clusters, the tested groups---and hence the hypothesis itself---depend on the observed data.
This dual use of the data for both cluster discovery and subsequent testing, often referred to as double dipping \citep{kriegeskorte2009circular}, induces selection bias and invalidates classical $p$-values, preventing reliable control of the Type~I error rate.

Selective inference (SI) provides a principled framework for addressing this problem by conducting statistical inference conditional on the data-dependent selection event \citep{taylor2015statistical,lee2016exact}.
Recent studies have developed SI methods for clustering, including hierarchical clustering \citep{gao2022selective} and $k$-means \citep{chen2023selective}, as well as feature-wise testing of differences between selected clusters \citep{chen2023testing}.
By explicitly accounting for the clustering process that generated the hypotheses, these approaches enable valid post-clustering inference without requiring data splitting.
Existing methods, however, primarily consider clustering algorithms that operate directly on the observed high-dimensional features.

Extending this framework to deep clustering poses a distinct technical challenge.
In sequential deep clustering, cluster assignments are determined not directly from the observed data but through the composition of a nonlinear encoder and a discrete clustering algorithm.
Even when the encoder is pretrained on an independent reference dataset and its parameters are fixed during inference, the latent representation---and, in particular, the activation pattern induced by a piecewise-affine neural network---still depends on the target data.
Consequently, the selection event associated with the final clustering is substantially more complex than that arising when clustering is applied directly to the observed features.
Constraints that characterize cluster assignments in the latent space are valid only within individual activation regions of the encoder and must therefore be appropriately combined across such regions.

In this work, we develop a selective-inference framework for sequential deep clustering with a fixed pretrained encoder.
The framework applies to fixed piecewise-affine encoders together with clustering procedures whose selection events can be characterized along affine perturbations of the latent representations; Multi-Layer Perceptron (MLP) encoders and $k$-means provide a representative instantiation developed in Section~\ref{sec:method} and used in our experiments.
We consider feature-wise inference for differences between clusters identified in the latent space, following the testing framework of \citet{chen2023testing}.
Our main methodological contribution is to characterize the selection event induced by the entire encoder-and-clustering pipeline in a computationally tractable form, thereby enabling valid statistical inference for feature differences between data-driven clusters.
We further develop a computational procedure that avoids excessive conditioning on internal algorithmic states and thereby improves statistical power while retaining selective validity.
Through synthetic experiments, we demonstrate that the proposed method controls the Type~I error rate where naive inference fails and achieves higher power than valid but conservative alternatives.
Applications to bulk and single-cell RNA-seq data further illustrate its utility for quantifying statistically supported feature differences between structures discovered by deep clustering.

\paragraph{Related Work}
\label{sec:related_work}

Selective inference (SI) provides a principled framework for valid statistical inference after data-dependent selection \citep{taylor2015statistical}.
Much of the modern SI literature was initially motivated by model and feature selection in regression, most notably the Lasso \citep{lee2016exact,fithian2014optimal,tibshirani2016exact}.
The central problem addressed by SI is selection bias caused by using the same data both to select a hypothesis and to test it, often referred to as double dipping.
Rather than ignoring the data-dependent selection process, SI conducts inference conditional on the selection event induced by the algorithm, thereby enabling valid statistical testing after selection.
Since these early developments, SI has been extended from classical feature-selection problems \citep{yang2016selective,suzumura2017selective,charkhi2018asymptotic,rugamer2020inference,sugiyama2021more,das2022fast,panigrahi2023approximate,shiraishi2025statistical} to increasingly complex data-dependent procedures and to settings beyond supervised variable selection.

SI is particularly useful for unsupervised learning, where the objects to be tested are themselves discovered from the data.
Examples include change points detected from sequential data \citep{duy2020computing,jewell2022testing,shiraishi2023selective}, outliers detected based on model deviations \citep{chen2020valid,tsukurimichi2022conditional}, and clusters discovered by clustering algorithms \citep{gao2022selective,chen2023selective,miyata2026clustering}.
In these problems, one first uses the data to identify a structure of interest and subsequently assesses its statistical significance using the same data, making selection bias inherent in the inferential task.
Unlike in supervised learning, standard data splitting is generally not directly applicable to such unsupervised inference problems, because the objects identified in one subset of the data do not necessarily have well-defined counterparts in another independent subset.
For clustering, for example, a partition discovered in one subset does not directly define the same clusters in another subset.
Recent studies have therefore developed SI for clustering performed directly on the observed features, including hierarchical clustering \citep{gao2022selective}, $k$-means \citep{chen2023selective}, and feature-wise testing of differences between selected clusters \citep{chen2023testing}, which we adopt in Section~\ref{sec:Preliminaries}.

A central practical challenge in SI is to characterize the selection event induced by the algorithm in a form that permits computation of the required conditional distribution.
This becomes substantially more difficult when the selection procedure involves deep neural networks.
Recent studies have developed SI for various data-dependent selection tasks involving deep neural networks, including neural network-based image segmentation \citep{duy2022quantifying}, saliency maps for convolutional and graph neural networks \citep{miwa2023salient,nishino2025statistical}, and attention maps in Vision Transformers \citep{shiraishi2024statistical,shiraishi2026statistical}.
Computational tools such as Auto-Conditioning \citep{miwa2023salient,katsuoka2025si4onnx} exploit the piecewise-affine structure of neural networks to characterize their selection events, while parametric programming \citep{le2021parametric} can mitigate the loss of power caused by over-conditioning.
Our work brings together these two lines of research---SI for clustering and SI for deep learning---to enable statistical inference for clusters discovered in a deep latent space.
Unlike existing deep-learning SI settings, where the downstream selection rule primarily involves thresholding or ranking network-derived quantities, sequential deep clustering combines a nonlinear encoder with a discrete clustering procedure whose internal updates---for example, the centroid and assignment updates of Lloyd's algorithm---depend on the data.
Consequently, valid inference requires characterizing the selection event induced jointly by the encoder and the clustering procedure.

\paragraph{Contributions}
\label{sec:contributions}

This work develops a statistical framework for conducting valid inference after deep clustering in a sequential setting.
Our contributions are as follows.

First, we rigorously characterize the selection event induced by the entire sequential deep clustering pipeline.
Unlike existing SI methods for clustering that operate directly on the observed features, our setting requires accounting for the composition of a piecewise-affine encoder and a subsequent clustering procedure.
By jointly characterizing the encoder's activation regions and the assignment decisions of the clustering algorithm, we obtain a tractable representation of the selection event and enable valid conditional inference for feature differences between clusters identified in the latent space.
As a concrete instantiation, Section~\ref{sec:method} develops this characterization for ReLU-based encoders and $k$-means (Lloyd's algorithm).

Second, we develop a computational procedure that mitigates the loss of statistical power caused by over-conditioning.
Building on the parametric-programming principle introduced for Lasso SI \citep{le2021parametric}, our procedure aggregates internal-state regions that yield the same final clustering output, thereby avoiding conditioning on a single algorithmic path while preserving selective validity.
This leads to higher statistical power than conservative single-region conditioning.

Finally, we demonstrate the effectiveness of our framework through experiments on both synthetic and real-world genomic datasets, including scRNA-seq and bulk RNA-seq profiles \citep{zhang2015comparison,zheng2017pbmc}.

\newpage
\section{Preliminaries}
\label{sec:Preliminaries}

In this section, we introduce the deep clustering used in this study and the statistical framework for feature-wise inference, highlighting the issue of selection bias.

\subsection{Deep Clustering}
\label{subsec:DeepClustering}

\begin{figure}[htbp]
\centering
\includegraphics[width=0.8\textwidth]{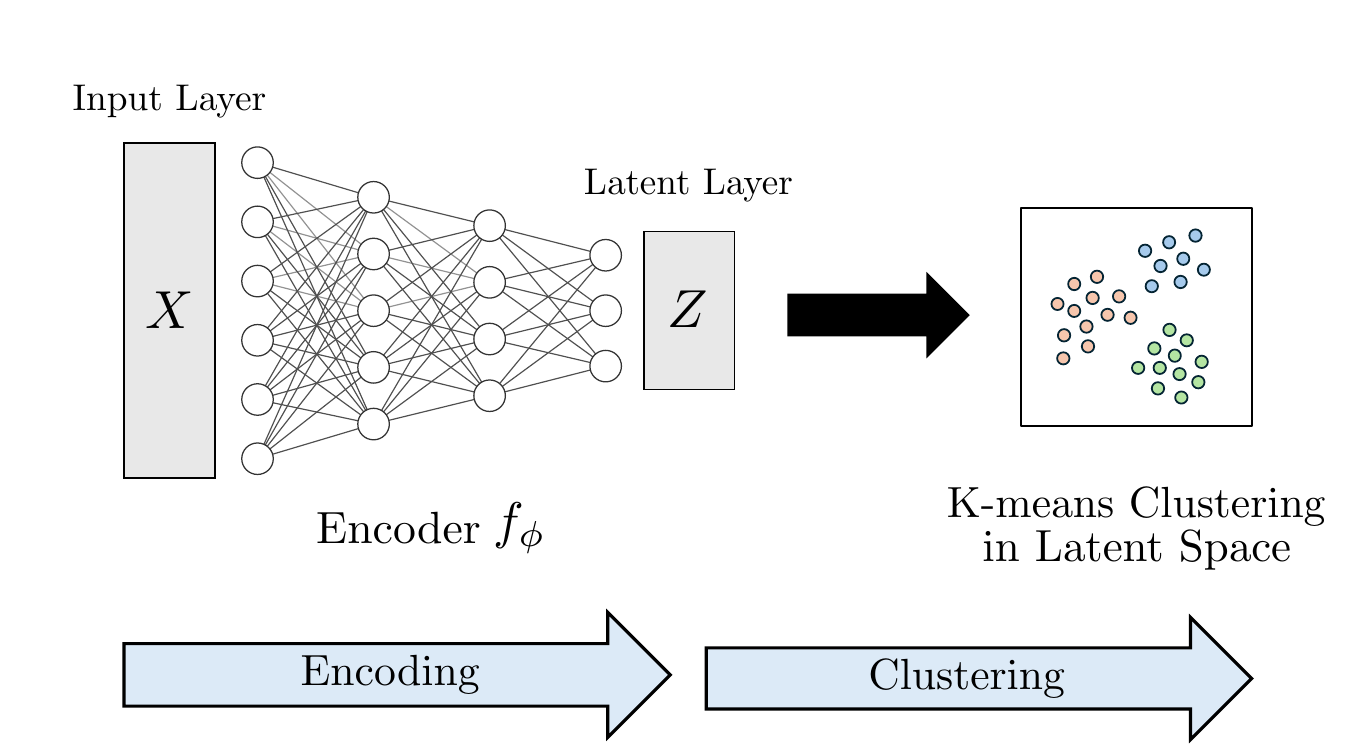}
\caption{Schematic illustration of sequential deep clustering with a fixed encoder followed by clustering in the latent space.}
\label{fig:DeepClustering}
\end{figure}

Clustering high-dimensional data directly can be challenging due to the curse of dimensionality \citep{beyer1999nearest}.
To address this issue, we adopt a sequential deep clustering approach, in which representation learning and clustering are performed as two separate stages \citep{zhou2025comprehensive}.
Mapping high-dimensional observations into a compact latent space before applying a conventional clustering algorithm can mitigate distance concentration and improve clustering performance when the learned representation preserves relevant structure \citep{geddes2019autoencoder}.
Let $X \in \mathbb{R}^{N \times D}$ be the data matrix with $N$ samples and $D$ features, where the $i$-th row vector is represented by $\bm{X}_i$.
Throughout the inference, the encoder parameters $\phi$ are treated as fixed: the encoder must be trained on data independent of the target dataset or otherwise supplied as an external fixed map. If the encoder is trained on the target data, its training procedure becomes part of the data-dependent selection event and is not covered by the present theory.
Our framework considers fixed piecewise-affine encoders together with clustering procedures whose selection events can be tractably characterized along affine perturbations of the latent representations (Figure~\ref{fig:DeepClustering}).

\subsection{Feature-wise Hypothesis Testing}
\label{subsec:HypothesisTesting}

For our probabilistic formulation, we regard the data matrix $X$ introduced above as a random matrix, and we write $\bm{X} \in \mathbb{R}^{ND}$ for its vectorization obtained by stacking the rows $\bm{X}_1, \dots, \bm{X}_N$; $X$ and $\bm{X}$ are equivalent representations of the same data.
We denote the observed realization by $\bm{x} \in \mathbb{R}^{ND}$, assume the following Gaussian observation model:
\begin{equation}
\bm{X} = \bm{\mu} + \bm{\varepsilon},
\quad
\bm{\varepsilon} \sim \mathcal{N}(\bm{0}, \bm{\Sigma}),
\label{eq:X_clustering}
\end{equation}
where $\bm{\mu} \in \mathbb{R}^{ND}$ is the unknown mean parameter and $\bm{\Sigma} \in \mathbb{R}^{ND\times ND}$ is a known positive-definite covariance matrix.
Let $M \in \mathbb{R}^{N\times D}$ denote the matrix form of $\bm{\mu}$ under the same row-stacking convention as $\bm{X}$, and write $\mu_{ij}:=M_{ij}$ for the mean of feature $j$ in sample $i$.
Equivalently, the data representation used for inference is assumed to follow $\bm{X}\sim\mathcal{N}(\bm{\mu},\bm{\Sigma})$.
For count-valued genomic observations, this assumption applies to the transformed data analyzed by the method, not to the raw counts.
Since $\bm{X}$ and $X$ represent the same data, we use $\mathcal{C}(\bm{X})$ to denote the same clustering outcome when working with the vectorized representation.

For a target feature $j$ and a pair of estimated clusters $\mathcal{C}_a$ and $\mathcal{C}_b$, we define the null hypothesis ($\mathrm{H}_{0,j}$) and the alternative hypothesis ($\mathrm{H}_{1,j}$) as follows:
\begin{equation}
\mathrm{H}_{0,j}:
\frac{1}{|\mathcal{C}_a|}\sum_{i\in\mathcal{C}_a}\mu_{ij}
=
\frac{1}{|\mathcal{C}_b|}\sum_{i\in\mathcal{C}_b}\mu_{ij}
\quad \text{vs.} \quad
\mathrm{H}_{1,j}:
\frac{1}{|\mathcal{C}_a|}\sum_{i\in\mathcal{C}_a}\mu_{ij}
\neq
\frac{1}{|\mathcal{C}_b|}\sum_{i\in\mathcal{C}_b}\mu_{ij}.
\label{eq:H0H1_clustering}
\end{equation}
We measure the separation between these clusters for feature $j$ using the difference in their empirical means.
The test statistic $T_j(\bm{X})$ is defined as:
\begin{equation}
T_j(\bm{X})
=
\frac{1}{|\mathcal{C}_a|}
\sum_{i\in\mathcal{C}_a}X_{ij}
-
\frac{1}{|\mathcal{C}_b|}
\sum_{i\in\mathcal{C}_b}X_{ij}
=
\bm{\eta}_j^\top \bm{X},
\label{eq:test_statistic_scalar}
\end{equation}
where $\bm{\eta}_j \in \mathbb{R}^{ND}$ is constructed to extract the difference in empirical means as a linear combination.
Specifically, letting $\bm{e}_j \in \mathbb{R}^D$ be the $j$-th standard basis vector and $\mathbb{I}(\cdot)$ be the indicator function, $\bm{\eta}_j$ is defined by concatenating $N$ block vectors $\bm{v}_1, \dots, \bm{v}_N \in \mathbb{R}^D$:
\begin{equation}
\bm{\eta}_j =
\begin{pmatrix}
\bm{v}_1 \\
\vdots \\
\bm{v}_N
\end{pmatrix},
\quad
\bm{v}_i =
\left(
\frac{\mathbb{I}(i \in \mathcal{C}_a)}{|\mathcal{C}_a|}
-
\frac{\mathbb{I}(i \in \mathcal{C}_b)}{|\mathcal{C}_b|}
\right) \bm{e}_j
\quad (i = 1, \dots, N).
\label{eq:eta_j_blocks}
\end{equation}
Because statistical hypothesis testing evaluates the hypotheses by computing $p$-values based on the sampling distribution of the test statistic, the core problem reduces to deriving the distribution of $T_j(\bm{X})$.

\subsection{Selection Bias}
\label{subsec:SelectionBias}
An unadjusted approach assumes that the test statistic follows a normal distribution:
\begin{equation}
T_j(\bm{X})
\sim
\mathcal{N}(\bm{\eta}_j^\top \bm{\mu},
\bm{\eta}_j^\top \bm{\Sigma} \bm{\eta}_j).
\label{eq:naive_T_normal}
\end{equation}

However, this inferential procedure is not valid in our context.
Because $\bm{\eta}_j$ is determined by cluster assignments derived from the same data $\bm{X}$, the resulting hypotheses are data-dependent.
This dual use of the data for both hypothesis generation and testing, a practice known as double dipping \citep{kriegeskorte2009circular}, induces selection bias and consequently fails to control the Type~I error rate.

\newpage
\section{Selective Inference for Deep Clustering}
\label{sec:method}

In this section, we develop a selective inference (SI) framework to yield valid selective $p$-values for deep clustering results.
SI corrects selection bias arising from data-dependent hypothesis selection by characterizing the non-linear selection event in a tractable form.
Recent advances have covered a range of deep models, from saliency maps of convolutional and graph neural networks \citep{miwa2023salient, nishino2025statistical} to change points detected by recurrent networks \citep{shiraishi2023selective} and anomalies flagged by variational autoencoders \citep{miwa2024statistical}.
Drawing on these developments, we extend the selective inference framework to clustering results obtained from deep latent features.
Following the standard selective-inference construction \citep{fithian2014optimal, lee2016exact}, we first condition on the observed clustering outcome and then eliminate nuisance parameters, thereby reducing the problem to inference along a one-dimensional search line.
In the remainder of this section, we illustrate the proposed framework using a piecewise-affine neural encoder and $k$-means clustering as a concrete example.
For the encoder, an MLP with ReLU activations provides a representative instance of the class of piecewise-affine neural networks.
We emphasize that the proposed framework is not intrinsically restricted to ReLU-based MLPs or $k$-means.
The same principle extends to other encoder architectures for which the induced latent representations can be tractably characterized along the selective-inference search line, and to other clustering procedures whose selection events can be characterized under the corresponding latent-space perturbations.

\subsection{Conditional Distribution and Selective $p$-value}

The test statistic $T_j(\bm{X})$, quantifying the difference in the $j$-th feature between clusters, is the linear contrast
\(
T_j(\bm{X}) = \bm{\eta}_j^\top \bm{X}
\)
in \eqref{eq:test_statistic_scalar}, with the block construction of $\bm{\eta}_j$ in \eqref{eq:eta_j_blocks} that encodes the selected cluster pair $(\mathcal{C}_a,\mathcal{C}_b)$ and the feature index~$j$.
To eliminate nuisance parameters, we further condition on the sufficient statistic $\bm{Q}(\bm{X})$ \citep{lee2016exact}:
\begin{equation}
\bm{Q}(\bm{X}) = \left( I_{ND} - \frac{\bm{\Sigma} \bm{\eta}_j \bm{\eta}_j^\top} {\bm{\eta}_j^\top \bm{\Sigma} \bm{\eta}_j} \right) \bm{X}.
\label{eq:sufficient_statistic}
\end{equation}
Conditioning on $\bm{Q}(\bm{X})=\bm{Q}(\bm{x})$ restricts the data to a one-dimensional search line
\begin{equation}
\bm{X}(z)=\bm{a}+\bm{b}z,
\quad z\in\mathbb{R},
\label{eq:affine_subspace}
\end{equation}
where $\bm{a} = \bm{Q}(\bm{x})$, $\bm{b} = \bm{\Sigma}\bm{\eta}_j / (\bm{\eta}_j^\top\bm{\Sigma}\bm{\eta}_j)$, and $z = T_j(\bm{X}(z))$ \citep{lee2016exact,chen2023selective}.
Along this line, the selection event $\mathcal{C}(\bm{X})=\mathcal{C}(\bm{x})$ corresponds to the truncation region
\begin{equation}
\mathcal{Z} = \{z \in \mathbb{R} \mid \mathcal{C}(\bm{X}(z)) = \mathcal{C}(\bm{x})\}.
\label{eq:feasible_region}
\end{equation}
The following theorem establishes the conditional distribution of $T_j(\bm{X})$ under this selection event.
It is the statement of \citet{chen2023selective} and \citet{chen2023testing}, unchanged: the derivation uses the selection event only through the truncation region $\mathcal{Z}$, so it applies verbatim once $\mathcal{Z}$ is available.
What is specific to deep clustering is therefore not the theorem but $\mathcal{Z}$ itself, whose characterization and computation occupy the remainder of this section.

\begin{theorem}
\label{thm:truncated_normal}
Consider $\bm{X} \sim \mathcal{N}(\bm{\mu}, \bm{\Sigma})$ and observed data $\bm{x}$.
Under the null hypothesis $\mathrm{H}_{0,j}$, conditional on $\mathcal{C}(\bm{X}) = \mathcal{C}(\bm{x})$ and $\bm{Q}(\bm{X}) = \bm{Q}(\bm{x})$, the conditional distribution of $T_j(\bm{X})$ is $\mathcal{TN}(0, \sigma_j^2, \mathcal{Z})$, where $\sigma_j^2 = \bm{\eta}_j^\top \bm{\Sigma} \bm{\eta}_j$ and $\mathcal{Z}$ is defined in \eqref{eq:feasible_region}.
\end{theorem}

Based on Theorem~\ref{thm:truncated_normal}, the selective $p$-value $p_{\mathrm{selective}}$ is obtained from the truncated normal law $\mathcal{TN}(0,\sigma_j^2,\mathcal{Z})$ in the usual way.
The following theorem guarantees the statistical validity of this $p$-value.
It is the standard selective-inference guarantee \citep{fithian2014optimal, lee2016exact} in the clustering form used by \citet{chen2023selective}; we restate it here because the truncation region $\mathcal{Z}$ entering the selective pivot is our own.

\begin{theorem}
\label{thm:valid_p}
The selective $p$-value $p_{\mathrm{selective}}$ satisfies
\begin{equation}
\mathbb{P}_{\mathrm{H}_{0,j}} \bigl( p_{\mathrm{selective}} \le \alpha \,\big|\, \mathcal{C}(\bm{X}) = \mathcal{C}(\bm{x}) \bigr) = \alpha,
\quad \forall \alpha \in (0,1).
\label{eq:valid_p_identity}
\end{equation}
\end{theorem}

Proofs are provided in Appendix~\ref{appendix:si_proofs}.
Thus $p_{\mathrm{selective}}$ can be evaluated once the truncation region $\mathcal{Z}$ is identified, which is the remaining computational task.

\subsection{Selection Region Identification and Problem Decomposition}
\label{subsec:decomposition}

The computational core of our method is interval identification along \eqref{eq:affine_subspace}: we seek the set of scalars $z$ for which the deep clustering applied to $\bm{X}(z)$ reproduces the observed clustering $\mathcal{C}(\bm{x})$, i.e., $\mathcal{Z}$ in \eqref{eq:feasible_region}.

To obtain algebraic constraints in the scalar $z$, we refine the selection event by the internal state $\mathcal{S}(\bm{X}) = \bigl( \mathcal{S}_{\mathrm{NN}}(\bm{X}), \mathcal{S}_{\mathrm{KM}}(\bm{X}) \bigr)$, comprising
\begin{itemize}
\item $\mathcal{S}_{\mathrm{NN}}(\bm{X})$, the encoder's activation pattern (which polyhedral region of the piecewise affine map contains $\bm{X}$), and
\item $\mathcal{S}_{\mathrm{KM}}(\bm{X})$, the symbolic path of $k$-means (the assignment decisions across iterations that determine the distance comparisons).
\end{itemize}
Fixing $\mathcal{S}(\bm{X})=s$ at any attainable internal state $s$ allows us to treat the encoder as affine in $\bm{X}$ and to express the $k$-means comparisons as polynomial inequalities in $z$.
For each attainable state $s$, define its fixed-state feasible set by
\begin{equation}
\mathcal{Z}_{\mathrm{poly}}^{(s)}
=
\{r \in \mathbb{R} \mid \mathcal{S}(\bm{X}(r))=s\}.
\label{eq:Z_poly_def}
\end{equation}
Because the final clustering is determined by the internal state, all points in $\mathcal{Z}_{\mathrm{poly}}^{(s)}$ have the same clustering output; in particular, $\mathcal{Z}_{\mathrm{poly}}^{(s)} \subseteq \mathcal{Z}$ whenever $s$ yields $\mathcal{C}(\bm{x})$.
The set \eqref{eq:Z_poly_def} need not be connected because the $k$-means constraints are quadratic.
For a current search point $z$, write $s_z=\mathcal{S}(\bm{X}(z))$ and define
\begin{equation}
I_z=[L_z,U_z]
\coloneqq
\text{the connected component of }\mathcal{Z}_{\mathrm{poly}}^{(s_z)}
\text{ that contains }z.
\label{eq:local_interval_def}
\end{equation}
We decompose the computation of this local interval into two sub-problems:
\begin{itemize}
\item \textbf{Problem A (Neural Network):} Find the maximal interval containing the current point $z$ on which the encoder's activation pattern is constant. Denote it by $[L_{\mathrm{NN}}, U_{\mathrm{NN}}]$.
\item \textbf{Problem B ($k$-means):} Find the maximal interval containing $z$ on which the cluster-assignment history is constant. Denote it by $[L_{\mathrm{KM}}, U_{\mathrm{KM}}]$.
\end{itemize}
The dependence of these two intervals on the current point $z$ is suppressed for readability.
Section~\ref{subsec:combine_AB} explains how their intersection yields $I_z$.

\subsection{Problem A: Neural Networks}

The first problem is to characterize the encoder along the search line.

\subsubsection{General piecewise affine structure}

Since the search line \eqref{eq:affine_subspace} is defined in the vectorized data space $\mathbb{R}^{ND}$, it is convenient to regard the encoder of Section~\ref{sec:Preliminaries} as acting on the whole dataset at once.
With a slight abuse of notation, we write
\begin{equation}
f_{\phi} : \mathbb{R}^{ND} \to \mathbb{R}^{Nd}, \qquad
\bm{X} \mapsto \bigl( f_{\phi}(\bm{X}_1)^\top, \ldots, f_{\phi}(\bm{X}_N)^\top \bigr)^\top,
\end{equation}
i.e., the same sample-wise encoder is applied to every row of $\bm{X}$.
The overall mapping $f_{\phi}$ is piecewise affine.
That is, the input space $\mathbb{R}^{ND}$ can be partitioned into a finite collection of convex polyhedra $\{\mathcal{R}_{\rho}\}_{\rho=1}^{R}$ such that
\begin{equation}
\forall \bm{X} \in \mathcal{R}_{\rho}, \quad
f_{\phi}(\bm{X}) = \bm{W}^{(\rho)} \bm{X} + \bm{c}^{(\rho)},
\label{eq:affine_map_general}
\end{equation}
for region-specific coefficients $\bm{W}^{(\rho)} \in \mathbb{R}^{Nd \times ND}$ and $\bm{c}^{(\rho)} \in \mathbb{R}^{Nd}$.
Because the encoder acts on each sample separately, $\bm{W}^{(\rho)}$ is block diagonal, its $i$-th block being the affine map induced by the activation pattern of sample $i$.

\subsubsection{Reduction along the search line}

Under selective inference, we restrict the input to the affine line $\bm{X}(z)=\bm{a}+\bm{b}z$ defined in \eqref{eq:affine_subspace}.
As long as this line stays inside a single region $\mathcal{R}_{\rho}$, the latent representation of the whole dataset is affine in $z$:
\begin{equation}
\bm{\zeta}(z) = f_{\phi}(\bm{X}(z)) = \bm{W}^{(\rho)} (\bm{a}+\bm{b}z) + \bm{c}^{(\rho)} = \tilde{\bm{a}} + \tilde{\bm{b}} z,
\label{eq:zeta_affine_z}
\end{equation}
where $\tilde{\bm{a}} = \bm{W}^{(\rho)}\bm{a} + \bm{c}^{(\rho)}$ and $\tilde{\bm{b}} = \bm{W}^{(\rho)}\bm{b}$ lie in $\mathbb{R}^{Nd}$.
Writing $\tilde{\bm{a}}_i, \tilde{\bm{b}}_i \in \mathbb{R}^d$ for the $i$-th blocks, the latent representation of sample $i$ is $\bm{\zeta}_i(z) = \tilde{\bm{a}}_i + \tilde{\bm{b}}_i z$; this is the quantity on which the $k$-means step of Problem~B operates.

Region boundaries are determined by changes in the activation pattern.
Let $\ell$ index the neurons of the encoder evaluated over all $N$ samples, and let the corresponding pre-activation along the line be written as
\[
u_{\ell}(z) = a_{\ell} + b_{\ell} z.
\]
The activation pattern remains unchanged as long as the sign of $u_{\ell}(z)$ does not change.
Thus, fixing the activation pattern induces a finite set of linear constraints of the form
\begin{equation}
a_{\ell} + b_{\ell} z \ge 0 \quad \text{or} \quad a_{\ell} + b_{\ell} z < 0.
\label{eq:linear_inequality}
\end{equation}

\paragraph{Geometric consequence}
Since these constraints are linear in $z$, the feasible set of $z$ that preserves the activation pattern is the intersection over all neurons $\ell$ of the regions defined by a single linear inequality.
On a one-dimensional line, the intersection of such semi-infinite intervals necessarily forms a single, connected interval $[L_{\mathrm{NN}}, U_{\mathrm{NN}}]$.
Moreover, because each neuron contributes at most one boundary point (given by $u_{\ell}(z)=0$), the total number of such points is finite.
This implies that the encoder induces a finite partition of the search line into distinct intervals.

\subsubsection{Implementation via Auto-Conditioning}

Explicitly enumerating all regions $\{\mathcal{R}_{\rho}\}$ is computationally infeasible.
Instead, we use Auto-Conditioning \citep{katsuoka2025si4onnx, miwa2024statistical}, which traces the computational graph of the encoder along the search line and propagates the linear constraints in \eqref{eq:linear_inequality} layer by layer.

Given any current search point $z$, this procedure computes the maximal interval $[L_{\mathrm{NN}}, U_{\mathrm{NN}}]$ containing $z$ such that the activation pattern at $\bm{X}(r)$ remains identical to that at $\bm{X}(z)$ for every $r \in [L_{\mathrm{NN}}, U_{\mathrm{NN}}]$.
At the observed realization $z=z_{\mathrm{obs}}=T_j(\bm{x})$, this is the interval that preserves the activation pattern at $\bm{x}$.

\subsection{Problem B: Piecewise Quadratic Constraints in $k$-means}
\label{subsec:problemB}

The $k$-means step assigns points using squared Euclidean distances in the latent space.
Fix a current search point $z$ and let $\mathcal{C}_k^{(t)}(z)$ be the index set assigned to cluster $k$ at iteration $t$ of the Lloyd run on $\bm{X}(z)$.
Using the affine map of the activation region containing $\bm{X}(z)$, extend the latent representation algebraically to a candidate coordinate $r$ and define the centroid obtained while holding these assignments fixed by
\[
\bm{m}_k^{(t)}(r;z)
=
\frac{1}{|\mathcal{C}_k^{(t)}(z)|}
\sum_{i \in \mathcal{C}_k^{(t)}(z)}
\bm{\zeta}_i(r).
\]
This affine extension agrees with the actual encoder output for $r\in[L_{\mathrm{NN}},U_{\mathrm{NN}}]$.
The centroid $\bm{m}_k^{(t)}(r;z)$ is affine in $r$, and each squared distance is a quadratic polynomial:
\begin{equation}
\|\bm{\zeta}_i(r) - \bm{m}_k^{(t)}(r;z)\|^2
=
A_{i,k}^{(t)}(z)r^2+B_{i,k}^{(t)}(z)r+C_{i,k}^{(t)}(z).
\label{eq:squared_distance}
\end{equation}
Writing $y_i^{(t+1)}(z)$ for the label assigned to point $i$ at the next assignment step of the anchor run, preservation of that decision requires
\[
\|\bm{\zeta}_i(r)-\bm{m}_{y_i^{(t+1)}(z)}^{(t)}(r;z)\|^2
\le
\|\bm{\zeta}_i(r)-\bm{m}_{k}^{(t)}(r;z)\|^2
\quad
\text{for every }k\ne y_i^{(t+1)}(z).
\]
Each such constraint is quadratic in the candidate coordinate $r$.
Since its solution set need not be an interval, we retain the connected component containing the anchor point $z$ while accumulating the constraints over all samples and Lloyd iterations.
This yields the maximal local interval $[L_{\mathrm{KM}}, U_{\mathrm{KM}}]$ containing the current point on which the symbolic $k$-means path is unchanged (Appendix~\ref{appendix:kmeans_constraints_z}).

\subsection{Combining Problems A and B}
\label{subsec:combine_AB}

At a current search point $z$, the local interval on which the entire internal state is unchanged is obtained by intersecting the two component intervals from Problems~A and~B:
\begin{equation}
I_z=[L_z,U_z]
=
[L_{\mathrm{NN}}, U_{\mathrm{NN}}]
\cap
[L_{\mathrm{KM}}, U_{\mathrm{KM}}].
\label{eq:local_interval_intersection}
\end{equation}
Equation~\eqref{eq:local_interval_intersection} is precisely the connected component of $\mathcal{Z}_{\mathrm{poly}}^{(s_z)}$ containing $z$, as defined in \eqref{eq:local_interval_def}; it does not assert that the whole fixed-state set $\mathcal{Z}_{\mathrm{poly}}^{(s_z)}$ is a single interval.
Other connected components of the same fixed-state set, as well as components associated with other internal states, are visited at later steps of the search in Section~\ref{subsec:parametric_pp}.

Since all constraints are defined by polynomial inequalities in $z$ (linear for the neural network and quadratic for $k$-means), the selection region $\mathcal{Z}$ is guaranteed to be a finite union of intervals.

\subsection{Parametric Programming}
\label{subsec:parametric_pp}

Although a single local interval $I_{z_{\mathrm{obs}}}$ is easy to compute, using only that interval as the truncation region would over-condition and reduce statistical power.
To avoid this loss, we construct the selection region $\mathcal{Z}$ by aggregating all local intervals whose clustering output agrees with $\mathcal{C}(\bm{x})$.

Following the parametric-programming construction of \citet{miyata2026clustering}, the local intervals along the search line are stitched into the global truncation set.
For each $z$, $I_z=[L_z,U_z]$ in \eqref{eq:local_interval_def} is the maximal interval containing $z$ on which the internal state, and hence the clustering output, coincides with that at $\bm{X}(z)$.
Therefore,
\begin{equation}
\mathcal{Z}
\;=\;
\bigcup_{\substack{z \in \mathbb{R} \\
\mathcal{C}(\bm{X}(z)) = \mathcal{C}(\bm{x})}}
I_z
\;=\;
\bigcup_{\substack{z \in \mathbb{R} \\
\mathcal{C}(\bm{X}(z)) = \mathcal{C}(\bm{x})}}
[L_z,U_z],
\label{eq:parametric_union_line}
\end{equation}
which is the clustering-only specialization of the construction used by \citet{miyata2026clustering}.
Equivalently,
\begin{equation}
\mathcal{Z}
\;=\;
\bigcup_{s \in \mathfrak{S}(\mathcal{C}(\bm{x}))}
\mathcal{Z}_{\mathrm{poly}}^{(s)},
\label{eq:Z_union_marginalization}
\end{equation}
where $\mathfrak{S}(\mathcal{C}(\bm{x}))$ is the set of internal states that yield the observed clustering and $\mathcal{Z}_{\mathrm{poly}}^{(s)}$ is defined in \eqref{eq:Z_poly_def}.
Equation~\eqref{eq:Z_union_marginalization} is equivalent to \eqref{eq:parametric_union_line} because the union of all connected components $I_z$ associated with a fixed state $s$ recovers the entire set $\mathcal{Z}_{\mathrm{poly}}^{(s)}$.

We dynamically identify $\mathcal{Z}$ using Algorithm~\ref{alg:parametric-si}, following the parametric-programming procedures used for deep neural networks in \citet{miwa2024statistical,katsuoka2025si4onnx}.
Let $\mathsf{DC}=(f_\phi,\Theta_{\mathrm{KM}})$ denote the fixed deep-clustering procedure, where $\Theta_{\mathrm{KM}}$ collects the number of clusters, finite iteration limit, initialization, tie-breaking, empty-cluster, and stopping rules specified in Section~\ref{subsec:DeepClustering}.
The procedure starts at $z_{\min}$ and traces the specified portion of the search line; at each current point $z$, it identifies $I_z=[L_z,U_z]$, the connected component of $\mathcal{Z}_{\mathrm{poly}}^{(s_z)}$ containing $z$.
If the resulting clustering matches $\mathcal{C}(\bm{x})$, this local interval is merged into $\mathcal{Z}$.
The search then moves to $U_z+\delta$, where $\delta$ is a small positive numerical offset used only to cross the current boundary, rather than a grid-search step.
Following \citet{miwa2024statistical}, the search bounds may be set to $z_{\min}=-|z_{\mathrm{obs}}|-10\sigma_j$ and $z_{\max}=|z_{\mathrm{obs}}|+10\sigma_j$, for which the omitted Gaussian tail probability is negligible.
As in prior implementations, Algorithm~\ref{alg:parametric-si} therefore computes the truncation region to numerical precision over this effective support; the exact validity result in Theorem~\ref{thm:valid_p} corresponds to the ideal full-line characterization in \eqref{eq:parametric_union_line}.

\begin{algorithm}[t]
    \caption{Parametric Selective Inference for Deep Clustering}
    \label{alg:parametric-si}
    \KwIn{observed data $\bm{x}$, covariance matrix $\bm{\Sigma}$, fixed procedure $\mathsf{DC}$, target feature $j$, target cluster-label pair $(k_a,k_b)$, search range $[z_{\min},z_{\max}]$, boundary offset $\delta$}
    \KwOut{selective $p$-value $p_{\mathrm{selective}}$}
    
    Run $\mathsf{DC}$ on $\bm{x}$ to obtain $\mathcal{C}_{\mathrm{obs}} \leftarrow \mathcal{C}(\bm{x})$\;
    Set $\mathcal{C}_a \leftarrow \{i \mid (\mathcal{C}_{\mathrm{obs}})_i=k_a\}$ and $\mathcal{C}_b \leftarrow \{i \mid (\mathcal{C}_{\mathrm{obs}})_i=k_b\}$\;
    Construct $\bm{\eta}_j$ from $(\mathcal{C}_a,\mathcal{C}_b,j)$ using \eqref{eq:eta_j_blocks}\;
    Compute $z_{\mathrm{obs}} \leftarrow T_j(\bm{x})=\bm{\eta}_j^\top\bm{x}$\;
    Compute variance $\sigma_j^2 \leftarrow \bm{\eta}_j^\top \bm{\Sigma} \bm{\eta}_j$\;
    Compute $\bm{a} \leftarrow \bm{Q}(\bm{x})$ and $\bm{b} \leftarrow \bm{\Sigma}\bm{\eta}_j/(\bm{\eta}_j^\top\bm{\Sigma}\bm{\eta}_j)$\;
    Verify that $z_{\mathrm{obs}}\in[z_{\min},z_{\max}]$\;
    Initialize $\mathcal{Z} \leftarrow \emptyset$ and $z \leftarrow z_{\min}$\;
    
    \While{$z < z_{\max}$}{
        Run $\mathsf{DC}$ on $\bm{X}(z)$ to obtain clustering $\mathcal{C}_{\mathrm{curr}}$ and internal state $\mathcal{S}_{\mathrm{curr}}$\;
        
        Compute the component intervals $[L_{\mathrm{NN}}, U_{\mathrm{NN}}]$ and $[L_{\mathrm{KM}}, U_{\mathrm{KM}}]$ containing $z$, and set $I_z=[L_z,U_z] \leftarrow [L_{\mathrm{NN}}, U_{\mathrm{NN}}] \cap [L_{\mathrm{KM}}, U_{\mathrm{KM}}]$\;
        
        \If{$\mathcal{C}_{\mathrm{curr}} = \mathcal{C}_{\mathrm{obs}}$}{
            $\mathcal{Z} \leftarrow \mathcal{Z} \cup (I_z \cap [z_{\min},z_{\max}])$\;
        }
        
        $z \leftarrow U_z + \delta$\;
    }
    
    Compute $p_{\mathrm{selective}}$ from $\mathcal{TN}(0,\sigma_j^2,\mathcal{Z})$ at $z_{\mathrm{obs}}$\;
\end{algorithm}

\clearpage
\section{Numerical Experiments}
\label{sec:Experiments}
\label{sec:ArtificialExperiment}

In this section, we evaluate the statistical properties of the proposed method using synthetic and real-world datasets.
We verify the validity of the test under the null hypothesis (Type~I error rate control) and its ability to detect genuine signals under the alternative hypothesis (statistical power).
Section~\ref{subsec:synthetic} and Section~\ref{subsec:real} present the results for synthetic and real data, respectively.
We also conducted several robustness experiments using synthetic data to evaluate the Type~I error rate control under non-ideal conditions, namely estimated variance and non-Gaussian noise (see Appendix~\ref{appendix:robustness} for details).
The experiment scripts, preprocessing code, and trained encoder models are publicly available at \url{https://github.com/mizuieinanagoyaml/si4ae-clustering}.

\subsection{Synthetic Data Experiments}
\label{subsec:synthetic}

\subsubsection{Methods for Comparison}
\label{subsec:comparison_methods}

We evaluate the performance of the \texttt{Proposed} method by comparing it against three baselines:
(i)~\texttt{naive}, a conventional test that utilizes unadjusted $p$-values, ignoring the selection bias inherent in the clustering process;
(ii)~\texttt{Bonferroni}, which applies a conservative correction for multiple comparisons over all $K^N$ potential cluster assignments as:
\begin{equation}
p_{\mathrm{Bonferroni}} = \min(1, K^N \cdot p_{\mathrm{naive}});
\end{equation}
and (iii)~\texttt{w/o-pp}, an ablation study that computes $p$-values based on a single truncation interval without using parametric programming. This baseline is included to demonstrate the power gain achieved by our proposed interval integration.

\subsubsection{Experimental Setup}
\label{subsec:ExpSettings}

We generated synthetic datasets according to the Gaussian observation model in Equation~\eqref{eq:X_clustering} with identity covariance $\bm{\Sigma} = I_{ND}$.
For each trial, we set the number of clusters to $K=3$ and randomly selected a target feature $j$ and a pair of clusters for testing.
The significance level was set to $\alpha = 0.05$.
We employed a Multi-Layer Perceptron (MLP) autoencoder with ReLU activations; specific architectures for varying input dimensions $D$ are summarized in Table~\ref{tab:architectures}.
The encoder weights were pre-trained to convergence and treated as fixed parameters during inference.
To ensure reproducibility, we fixed random seeds for data generation and $k$-means initialization, and ran $k$-means for at most 300 iterations with convergence tolerance $10^{-4}$.

We adopted a symmetric decoder for each encoder and set the latent dimension to $d=128$.
We placed a ReLU activation after each fully connected layer (except the latent and output layers), used He initialization, and trained by minimizing mean squared reconstruction error until convergence.

\begin{table}[htbp]
    \centering
    \caption{Encoder architectures used in the synthetic data experiments}
    \label{tab:architectures}
    \begin{tabular}{llc}
        \hline
        Input Dimension $D$ & Encoder Layers (Number of Nodes) & Latent Dimension $d$ \\ \hline
        250   & $D \to 192 \to 128$ & 128 \\
        500   & $D \to 256 \to 128$ & 128 \\
        750   & $D \to 512 \to 256 \to 128$ & 128 \\
        1000  & $D \to 512 \to 256 \to 128$ & 128 \\ \hline
    \end{tabular}
\end{table}

\paragraph{Type~I error rate experiments}
To evaluate the validity of the proposed method, we examined the Type~I error rate using null datasets with no cluster structure ($\bm{\mu} = \bm{0}$). Following the Gaussian observation model in Eq.~\eqref{eq:X_clustering}, we conducted experiments under the following two scenarios:
\begin{itemize}
    \item Varying sample size $N$: $N \in \{100, 200, 300, 400, 500\}$ with $D=500$.
    \item Varying dimension $D$: $D \in \{250, 500, 750, 1000\}$ with $N=200$.
\end{itemize}
In each scenario, the empirical error rate was calculated over 10,000 independent trials at the significance level $\alpha=0.05$.

\paragraph{Power experiments}
To assess the statistical power, we evaluated the statistical power using datasets with a clear three-cluster structure ($K=3$). The cluster centers were set as detailed in Appendix~\ref{appendix:DataGeneration}. The experimental setup was as follows:
\begin{itemize}
    \item Varying separation: The distance between centroids was controlled by $\mathrm{sep} \in \{10, 20, 30, 40, 50, 60\}$.
    \item Fixed parameters: We fixed the data size to $(N, D) = (200, 500)$.
    \item Data generation: For each cluster, we generated $N/3$ samples by adding i.i.d.\ noise $\bm{\varepsilon} \sim \mathcal{N}(\bm{0}, \bm{I}_D)$ to the centroids.
\end{itemize}
Statistical power was estimated as the rejection rate of $\mathrm{H}_{0,j}$ over 10,000 independent trials.

\subsubsection{Results}
\label{subsec:results}

\begin{figure}[htbp]
    \centering
    \begin{subfigure}{0.45\textwidth}
        \centering
        \includegraphics[width=\textwidth]{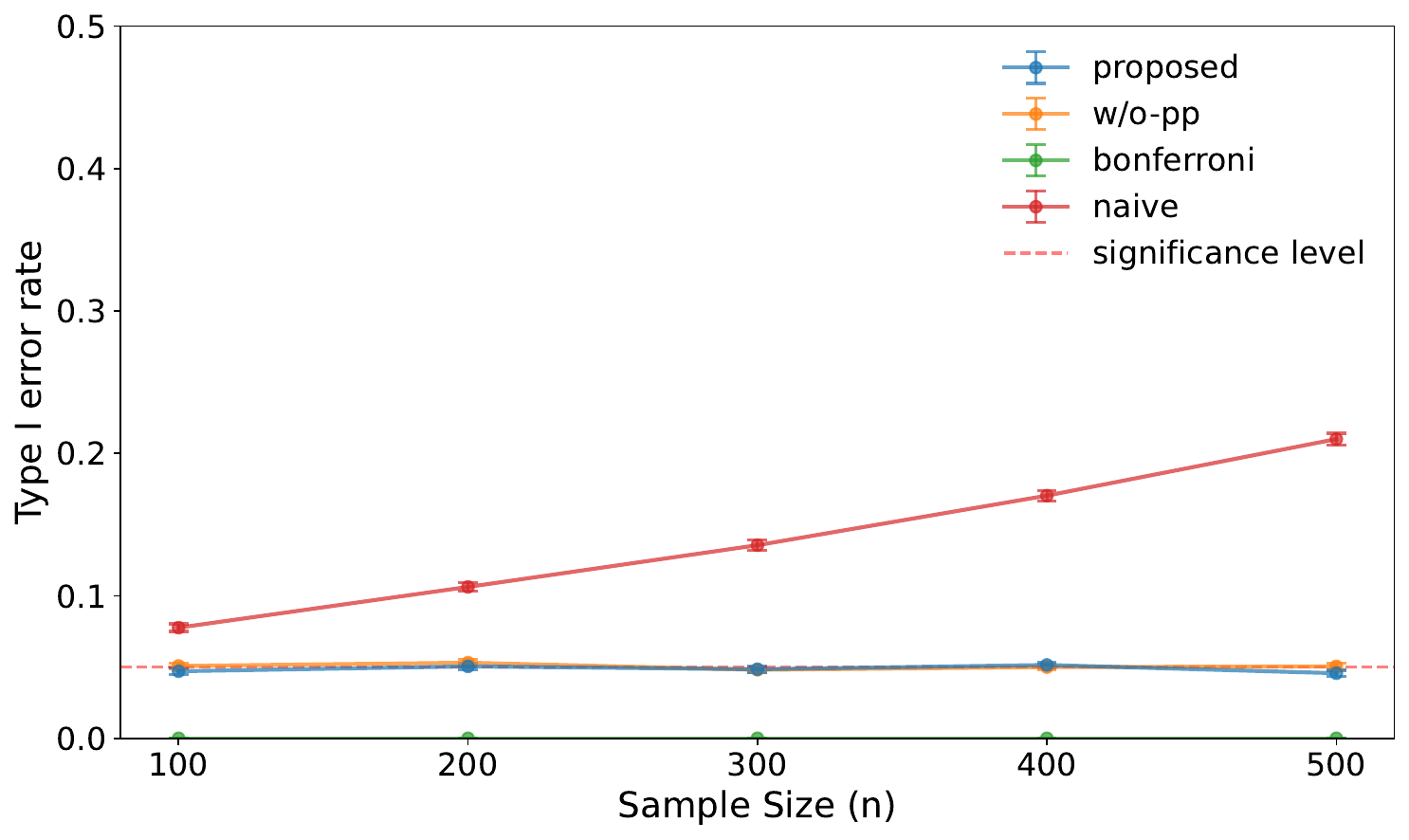}
        \caption{Type~I error rate against changes in $N$}
        \label{fig:fpr_n}
    \end{subfigure}
    \hfill
    \begin{subfigure}{0.45\textwidth}
        \centering
        \includegraphics[width=\textwidth]{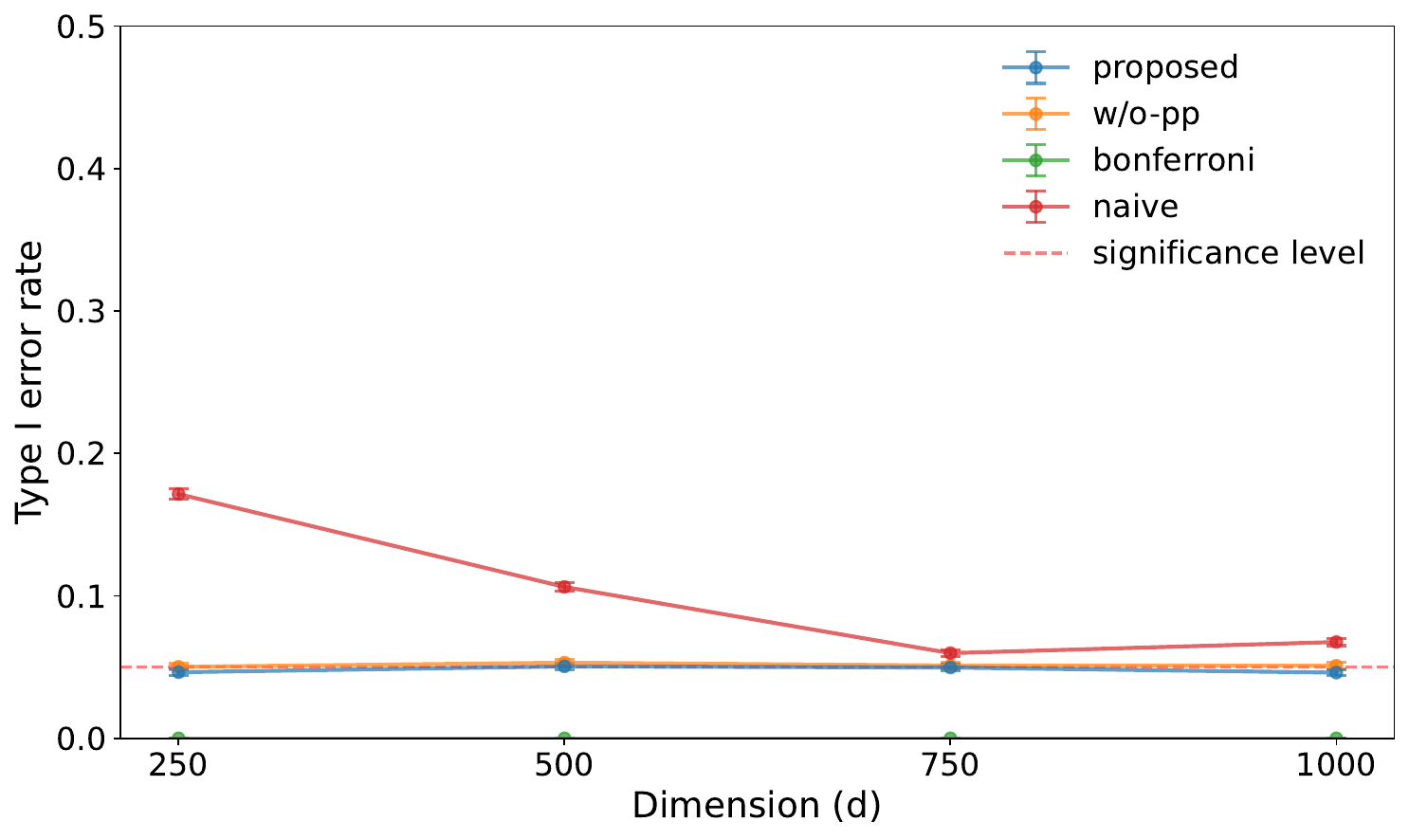}
        \caption{Type~I error rate against changes in $D$}
        \label{fig:fpr_d}
    \end{subfigure}
    \caption{Results of the Type~I error rate experiments}
    \label{fig:type_i_error}
\end{figure}

\begin{figure}[htbp]
    \centering
    \includegraphics[width=0.45\textwidth]{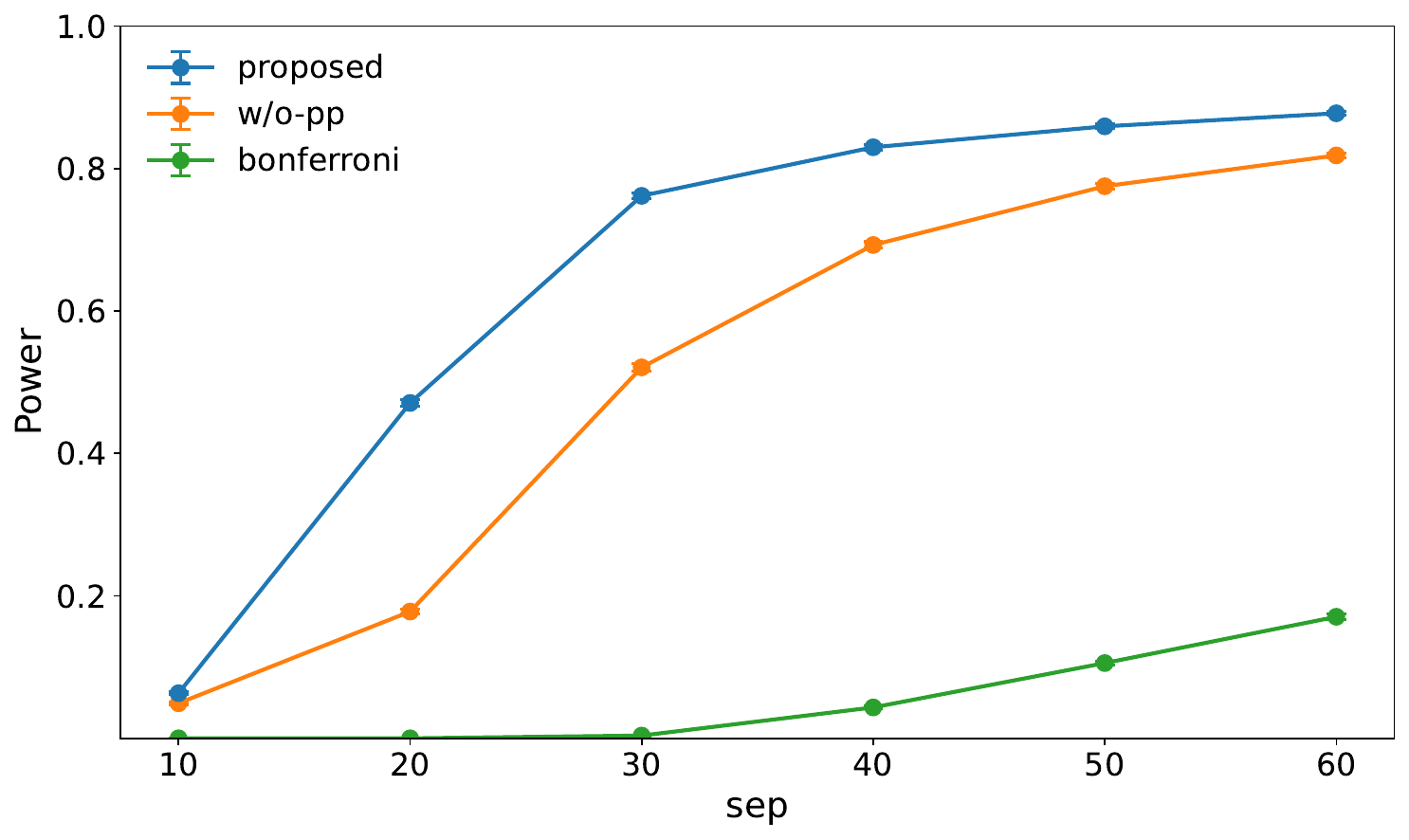}
    \caption{Results of the power experiments}
    \label{fig:power}
\end{figure}

Figures~\ref{fig:type_i_error} and~\ref{fig:power} summarize the synthetic experiments.

\paragraph{Type~I error rate}
The \texttt{naive} method fails to control the Type~I error rate.
Its empirical error rate exceeds the significance level $\alpha = 0.05$ across all settings.
This confirms that selection bias caused by clustering leads to an increased number of false discoveries.
In contrast, \texttt{Proposed}, \texttt{w/o-pp}, and \texttt{Bonferroni} maintain the error rate below $\alpha$ for all choices of $N$ and $D$.
These results are consistent with the selective validity of our framework for latent spaces formed by fixed deep neural networks.

\paragraph{Statistical power}
Among the methods that successfully controlled the Type~I error rate, \texttt{Proposed} consistently achieves the highest power.
\texttt{Bonferroni} is overly conservative due to the extremely large number of potential hypotheses ($K^N$).
Similarly, \texttt{w/o-pp} loses power because it is over-conditioned on a single truncation interval.
The proposed method achieves substantially higher detection ability by using parametric programming to aggregate multiple compatible intervals.

\clearpage
\subsection{Real Data Experiments}
\label{subsec:real}

We evaluate the proposed procedure on two gene-expression benchmarks: bulk RNA-seq of neuroblastoma (SEQC) and single-cell PBMC profiles with cell-type annotations.
We summarize how learned clusters align with available labels and compare testing methods by the number of significant genes at level $\alpha$.
The reported counts use feature-wise selective $p$-values and do not by themselves control family-wise error or false discovery rate across genes, cluster pairs, or choices of $K$.
These analyses are descriptive applications rather than evaluations of Type~I error or power, because the gene-level null hypotheses are unknown and transformed RNA-seq data need not satisfy the Gaussian model exactly. Statistical significance also does not inherently imply biological or scientific significance.
Extended biological context and gene-level discussions are deferred to Appendix~\ref{appendix:realdata_biology}.

\subsubsection{Analysis on the SEQC (Neuroblastoma) Dataset}
\label{subsec:SEQC_Experiment}

We utilized the SEQC neuroblastoma dataset (GEO: GSE49711), which contains bulk RNA-seq profiles of neuroblastoma samples \citep{zhang2015comparison}.
Importantly, because this is bulk rather than single-cell data, the observed profiles are influenced by multiple confounding factors, such as tumor purity and normal tissue admixture.
Therefore, rather than strictly defining the resulting clusters as distinct biological subtypes, we interpret them as major data-driven variation structures to evaluate the behavior of our proposed test.

\paragraph{Experimental Setup}

After applying a $\log_2(x + 1)$ transformation, we selected the top 100 highly variable genes ($D = 100$).
We evaluated two scenarios:
\begin{itemize}
    \item[] \textbf{1) Low-risk-only:} Restricted to 322 low-risk samples (split into 222 training and 100 test samples).
    \item[] \textbf{2) Risk-mixed:} Comprising all 498 samples, which include a mixture of both high-risk and low-risk patients (split into 398 training and 100 test samples).
\end{itemize}
In both settings, we standardized the test data via Z-score normalization using the training mean and variance.
For representation learning, we trained an MLP-based autoencoder using ReLU activations.
The encoder has an architecture of $D \rightarrow 64 \rightarrow 32 \rightarrow 8$ (latent dimension $d = 8$), and the decoder has a symmetric structure ($8 \rightarrow 32 \rightarrow 64 \rightarrow D$).
The models were trained individually for each setting by minimizing the mean squared error loss until convergence on the training split.

During the inference phase, we fixed the weights of the trained encoder and applied deep clustering to the test split for $K \in \{2, 3, 4\}$.
Finally, for each cluster pair, we computed the selective $p$-values for all 100 genes in the input space.

\paragraph{Discussion of Results}
We first examined whether the learned clusters were consistent with available clinical labels. As detailed in Appendix~\ref{appendix:cluster_details}, the resulting clusters broadly overlap with clinical risk groups in an unsupervised setting (see Figure~\ref{fig:tsne_appendix} for t-SNE visualizations). This overlap suggests that the learned latent space captures structure that is at least partially aligned with known risk stratification, providing a plausible setting for subsequent feature-wise inference.
Next, we compare testing methods by the number of significant genes detected by each procedure (Table~\ref{tab:seqc_results}).

\paragraph{Low-risk-only Setting}
The Low-risk-only setting serves as a baseline to evaluate how each method responds to a dataset that lacks clear clinical differences. However, because this is bulk RNA-seq data, the model may still detect subtle internal variations even without risk-based labels.
In the $K=2$ case, the \texttt{Proposed} method identifies 22 significant genes, consistent with feature-wise differences between the two selected clusters after adjustment for selection. As $K$ increases to 3 and 4, the partitions become more granular and the number of rejections by the \texttt{Proposed} method decreases sharply. Because the true feature-wise null hypotheses are unknown in these data, these counts alone do not establish robustness to over-clustering; they suggest that the proposed adjustment is substantially less prone than the naive analysis to declaring widespread differences in the finer partitions.
In contrast, the \texttt{naive} method consistently yields high rejection rates across all values of $K$. This persistent inflation of the significant-gene count is consistent with contamination by artificial differences created by the clustering process itself.

\paragraph{Comparison across Settings}
While the low rejection counts of the \texttt{w/o-pp} method in the Low-risk-only setting might initially appear reassuring, its behavior in the Risk-mixed setting points to a substantial loss of sensitivity.
In the Risk-mixed setting, where clusters tend to overlap with clinical risk, \texttt{w/o-pp} remains highly restrictive and identifies few significant genes despite apparently clearer between-cluster differences. In contrast, the \texttt{Proposed} method yields many more rejections. For instance, in the $K=2$ Risk-mixed case, it identifies 47 genes, matching the \texttt{naive} count while accounting for the clustering selection event.
Across all settings, the \texttt{Proposed} method consistently identifies more genes than \texttt{w/o-pp} and fewer than or the same number as the \texttt{naive} method. By using parametric programming to combine compatible truncation intervals, it is less conservative than single-interval conditioning without reverting to unadjusted testing. Because the true null status of each gene is unknown in these real data, the rejection counts cannot by themselves be interpreted as power or false discovery rates; they instead illustrate the practical difference among the three inferential procedures.

\begin{table}[htbp]
    \centering
    \caption{Proportion of significant genes detected in the SEQC dataset (number of significant genes / total 100 genes)}
    \label{tab:seqc_results}
    \begin{subtable}{\textwidth}
        \centering
        \caption{Low-risk-only setting}
        \begin{tabular}{ccccc}
            \hline
            $K$ & Cluster pair & naive & w/o-pp & proposed \\
            \hline
            2 & 0-1 & 31/100 & 3/100 & 22/100 \\
            \hline
            3 & 0-1 & 34/100 & 0/100 & 4/100 \\
            3 & 0-2 & 17/100 & 0/100 & 2/100 \\
            3 & 1-2 & 13/100 & 0/100 & 1/100 \\
            \hline
            4 & 0-1 & 21/100 & 0/100 & 2/100 \\
            4 & 0-2 & 10/100 & 0/100 & 1/100 \\
            4 & 0-3 & 12/100 & 0/100 & 0/100 \\
            4 & 1-2 & 37/100 & 0/100 & 6/100 \\
            4 & 1-3 & 17/100 & 0/100 & 0/100 \\
            4 & 2-3 & 29/100 & 0/100 & 0/100 \\
            \hline
        \end{tabular}
    \end{subtable}
    \vspace{3mm}
    \begin{subtable}{\textwidth}
    \centering
        \caption{Risk-mixed setting}
        \begin{tabular}{ccccc}
            \hline
            $K$ & Cluster pair & naive & w/o-pp & proposed \\
            \hline
            2 & 0-1 & 47/100 & 26/100 & 47/100 \\
            \hline
            3 & 0-1 & 58/100 & 3/100 & 24/100 \\
            3 & 0-2 & 53/100 & 2/100 & 24/100 \\
            3 & 1-2 & 34/100 & 1/100 & 10/100 \\
            \hline
            4 & 0-1 & 27/100 & 2/100 & 8/100 \\
            4 & 0-2 & 48/100 & 7/100 & 14/100 \\
            4 & 0-3 & 49/100 & 9/100 & 12/100 \\
            4 & 1-2 & 44/100 & 5/100 & 20/100 \\
            4 & 1-3 & 42/100 & 6/100 & 28/100 \\
            4 & 2-3 & 31/100 & 1/100 & 7/100 \\
            \hline
        \end{tabular}
    \end{subtable}
\end{table}

\subsubsection{Analysis on the PBMC (Immune Cells) Dataset}
\label{subsec:PBMC_Experiment}

We further applied our method to a PBMC (Peripheral Blood Mononuclear Cell) dataset provided by 10x Genomics \citep{zheng2017pbmc}.
Unlike the SEQC dataset, the PBMC data include cell-type annotations. These labels allow us to assess whether the learned clusters are consistent with known cell populations, but they do not reveal the true null status of individual gene contrasts and therefore cannot verify false-discovery control.

\paragraph{Experimental Setup}

After performing quality control via the Median Absolute Deviation (MAD) criterion, we applied library-size normalization and a $\log(1 + x)$ transformation to the raw counts.
We selected the top 1,000 highly variable genes based on the training set ($D = 1000$).
The features were then standardized using the training mean and variance.
We evaluated the following two settings:
\begin{itemize}
    \item[] \textbf{1) Single-type setting:} Restricted to Memory T cells to simulate a scenario without clear cluster structures ($n_{\mathrm{train}} = 8{,}237$, $n_{\mathrm{test}} = 1{,}500$).
This setting examines how each method behaves when clear between-type structure is absent, while residual within-type variation may still be present.
    \item[] \textbf{2) Multi-type setting:} Comprising Memory T, Natural Killer, and Naive Cytotoxic T cells to represent a structured population ($n_{\mathrm{train}} = 21{,}000$, $n_{\mathrm{test}} = 1{,}500$). This setting examines detection behavior across distinct lymphocyte lineages.
\end{itemize}
For representation learning, we employed an MLP-based autoencoder with an architecture of $D \rightarrow 512 \rightarrow 128 \rightarrow 64$ (latent dimension $d = 64$).
The decoder followed a symmetric structure.
We applied batch normalization and ReLU activations after each fully connected layer, except for the latent and final output layers, which used linear transformations. At inference time, batch normalization was run in evaluation mode with its learned running means and variances fixed, so each such layer remained affine.
The models were trained using He initialization until the reconstruction loss converged on the training split.
During the inference phase, we fixed the encoder weights and applied deep clustering to the test split with $K=3$.
We then computed selective $p$-values for all 1,000 genes across all cluster pairs.

\paragraph{Discussion of Results}
We first examined whether the deep clustering partitions were consistent with available cell-type annotations. In the Multi-type setting, the $K=3$ clusters largely overlap with the annotated cell types: Cluster~0 corresponds primarily to Naive Cytotoxic T cells, Cluster~1 to Natural Killer cells, and Cluster~2 to Memory T cells. This consistency suggests that the learned representation captures lineage-related structure that is useful for interpreting subsequent feature-wise tests, although annotation agreement alone does not establish the truth of individual gene-level hypotheses. Detailed t-SNE visualizations and a quantitative clustering summary are provided in Appendix~\ref{appendix:realdata_biology}.
Next, we compare the number of significant genes detected by each method (Table~\ref{tab:pbmc_results}).

\paragraph{Single-type Setting}
In the Single-type setting, the \texttt{naive} method rejected the null hypothesis for more than half of the genes in most cluster pairs. While intrinsic heterogeneity, such as differences in cell cycle or activation states, may exist even within a single cell type, such an extreme rejection rate is difficult to reconcile with the expected relative homogeneity of the cell population and is consistent with substantial selection bias. In contrast, the \texttt{Proposed} method and \texttt{w/o-pp} yielded substantially lower rejection counts, suggesting that both procedures mitigate clustering-induced bias relative to the unadjusted analysis.

\paragraph{Comparison across Settings}
In the Multi-type setting, all methods showed a marked increase in rejections, consistent with stronger between-lineage differences than in the Single-type setting. The \texttt{Proposed} method consistently identified more genes than \texttt{w/o-pp} across all cluster pairs. For instance, in the $0$--$1$ pair, the \texttt{Proposed} method detected 639 significant genes, whereas \texttt{w/o-pp} detected 436. This gap is consistent with the expected conservatism of conditioning on a single internal-state interval. By aggregating compatible truncation intervals via parametric programming, the \texttt{Proposed} method recovers additional rejections while retaining the selection adjustment. These findings illustrate a more favorable sensitivity trade-off than single-interval conditioning in this application, although the absence of gene-level ground truth precludes a direct estimate of power or false-discovery rates. Representative gene $p$-values and biological notes appear in Appendix~\ref{appendix:realdata_biology}.

\begin{table}[htbp]
    \centering
    \caption{Proportion of significant genes detected in the PBMC dataset (number of significant genes / total 1,000 genes)}
    \label{tab:pbmc_results}
    \begin{tabular}{lcccc}
        \hline
        Setting & Cluster Pair & naive & w/o-pp & proposed \\
        \hline
        Single-type & 0-1 & 507/1000 & 5/1000 & 34/1000 \\
        Single-type & 0-2 & 426/1000 & 26/1000 & 99/1000 \\
        Single-type & 1-2 & 717/1000 & 6/1000 & 78/1000 \\
        \hline
        Multi-type  & 0-1 & 668/1000 & 436/1000 & 639/1000 \\
        Multi-type & 0-2 & 452/1000 & 208/1000 & 279/1000 \\
        Multi-type  & 1-2 & 732/1000 & 656/1000 & 699/1000 \\
        \hline
    \end{tabular}
\end{table}

\clearpage
\section{Conclusions and Future Work}
\label{sec:conclusion}

In this work, we developed a selective-inference framework for statistically characterizing clusters identified in a learned latent space.
By accounting for the data-dependent selection process induced jointly by a fixed encoder and a subsequent clustering procedure, the proposed framework enables valid feature-wise inference for differences between selected clusters.
We further introduced a computational strategy based on parametric programming that aggregates compatible selection regions, thereby avoiding unnecessary over-conditioning.
Synthetic experiments demonstrated Type~I error control and improved statistical power over conservative alternatives, while applications to bulk and single-cell RNA-seq data illustrated the practical utility of the framework for identifying features that characterize data-driven subgroups.

Several directions remain for future work.
First, although the present methodological development focuses on piecewise-affine neural encoders and $k$-means clustering as a concrete realization, the underlying framework may be extended to broader classes of encoders and clustering algorithms for which the corresponding selection events can be characterized tractably.
Second, extending the theory to settings in which the encoder is trained on the same data used for inference is an important challenge, since the training procedure must then be incorporated into the selection event.
Other important directions include relaxing the Gaussian and known-covariance assumptions, developing simultaneous inference procedures that account for multiple features and cluster comparisons, and improving computational scalability for larger datasets and more complex deep clustering pipelines.

\newpage
\subsubsection*{Acknowledgments}
This work was partially supported by JST CREST (JPMJCR21D3, JPMJCR22N2), JST Moonshot R\&D (JPMJMS2033-05), and RIKEN Center for Advanced Intelligence Project.

\clearpage
\appendix

\newpage
\section{Proofs}
\label{appendix:si_proofs}

This appendix provides derivations supporting Theorems~\ref{thm:truncated_normal} and~\ref{thm:valid_p} in the main text.
Neither theorem is new: both are the corresponding results of \citet{chen2023selective} and \citet{chen2023testing}, and we reproduce their arguments here only to make explicit that they depend on the selection event solely through the truncation region $\mathcal{Z}$, and hence transfer unchanged from clustering in the original space to clustering in a learned latent space.
The first part follows the standard selective-inference search-line construction of \citet{lee2016exact}.
The second part records a conditioning/marginalization argument for the selective $p$-value.
The third part explains how refining the selection event by an internal state $\mathcal{S}$ and aggregating compatible intervals via parametric programming relates to computing the truncation region $\mathcal{Z}$ without over-conditioning.

\subsection{Proof of Theorem~\ref{thm:truncated_normal}}

Conditioning on $\bm{Q}(\bm{X}) = \bm{Q}(\bm{x})$ restricts $\bm{X}$ to the affine line $\bm{X}(z) = \bm{a} + \bm{b} z$ in Equation~\eqref{eq:affine_subspace}.
Imposing $\mathcal{C}(\bm{X}) = \mathcal{C}(\bm{x})$ further restricts $z$ to the truncation region $\mathcal{Z}$ defined in Equation~\eqref{eq:feasible_region}.
Along this line, $T_j(\bm{X}(z)) = z$.
Since $\bm{X} \sim \mathcal{N}(\bm{\mu}, \bm{\Sigma})$, the scalar $z = T_j(\bm{X})$ is marginally Gaussian with mean $\bm{\eta}_j^\top \bm{\mu}$ and variance $\sigma_j^2 = \bm{\eta}_j^\top \bm{\Sigma} \bm{\eta}_j$.
Moreover,
\[
\operatorname{Cov}\!\left(\bm{Q}(\bm{X}),T_j(\bm{X})\right)
=
\left(I_{ND}-\frac{\bm{\Sigma}\bm{\eta}_j\bm{\eta}_j^\top}{\sigma_j^2}\right)
\bm{\Sigma}\bm{\eta}_j
=\bm{0}.
\]
Joint Gaussianity therefore implies that $\bm{Q}(\bm{X})$ and $T_j(\bm{X})$ are independent.
Under $\mathrm{H}_{0,j}$, we have $\bm{\eta}_j^\top \bm{\mu} = 0$, so the unrestricted law of $z$ is $\mathcal{N}(0, \sigma_j^2)$.
Conditioning on $\bm{Q}(\bm{X})=\bm{Q}(\bm{x})$ and restricting $z$ to $\mathcal{Z}$ therefore yields the truncated normal law $\mathcal{TN}(0, \sigma_j^2, \mathcal{Z})$.

\subsection{Proof of Theorem~\ref{thm:valid_p}}

We use the standard marginalization-over-nuisance argument \citep{fithian2014optimal, lee2016exact}, instantiated as in \citet{chen2023selective} with our clustering event and $\bm{Q}(\bm{X})$.
Let $\mathcal{A} = \{ \mathcal{C}(\bm{X}) = \mathcal{C}(\bm{x}) \}$.
Under $\mathrm{H}_{0,j}$, conditional on $\mathcal{A}$ and $\bm{Q}(\bm{X})=\bm{Q}(\bm{x})$, Theorem~\ref{thm:truncated_normal} implies that the selective pivot is uniform on $(0,1)$; consequently $p_{\mathrm{selective}}$ has the usual uniform selective property given $(\mathcal{A},\bm{Q}(\bm{X}))$.
Hence,
\[
\mathbb{P}_{\mathrm{H}_{0,j}}\bigl(p_{\mathrm{selective}}\le \alpha \,\big|\, \mathcal{A},\, \bm{Q}(\bm{X})=\bm{Q}(\bm{x})\bigr)=\alpha ,
\quad \forall \alpha\in(0,1).
\]
Applying the tower property for regular conditional probabilities while holding $\mathcal{A}$ fixed yields
\begin{align*}
\mathbb{P}_{\mathrm{H}_{0,j}}\bigl(p_{\mathrm{selective}}\le \alpha \,\big|\, \mathcal{A}\bigr)
&=
\mathbb{E}_{\mathrm{H}_{0,j}}\!\left[
\mathbb{P}_{\mathrm{H}_{0,j}}\bigl(p_{\mathrm{selective}}\le \alpha
\,\big|\,\mathcal{A},\bm{Q}(\bm{X})\bigr)
\,\middle|\,\mathcal{A}
\right]
=
\alpha ,
\end{align*}
which is \eqref{eq:valid_p_identity}.

\subsection{Internal states and parametric programming}

Theorem~\ref{thm:truncated_normal} and the selective $p$-value condition on the clustering event $\mathcal{A}$ and $\bm{Q}(\bm{X})$, but not on the internal algorithmic path $\mathcal{S}(\bm{X})$.
In Section~\ref{subsec:decomposition}, $\mathcal{S}$ is introduced only as a computational device: fixing $\mathcal{S}(\bm{X})=s$ yields the polynomially described fixed-state set $\mathcal{Z}_{\mathrm{poly}}^{(s)}$ in Equation~\eqref{eq:Z_poly_def}.
Under a fixed internal state, neural-network constraints are linear in $z$ and $k$-means constraints are quadratic in $z$, so $\mathcal{Z}_{\mathrm{poly}}^{(s)}$ is in general a finite union of intervals rather than a single interval.
At a current search point $z$, the intersection in Equation~\eqref{eq:local_interval_intersection} computes only $I_z=[L_z,U_z]$, the connected component of $\mathcal{Z}_{\mathrm{poly}}^{(s_z)}$ that contains $z$.

Conditioning permanently on a single observed path $\mathcal{S}(\bm{x})$ would discard other paths compatible with the same $\mathcal{C}(\bm{x})$ and introduce over-conditioning relative to $\mathcal{A}$.
Parametric programming aggregates the local intervals $I_z$ whose clustering output agrees with $\mathcal{C}(\bm{x})$, thereby constructing the union truncation $\mathcal{Z}$ targeted by Theorem~\ref{thm:truncated_normal}.
If $p_{\mathrm{selective}}$ is evaluated with this union truncation, the conditional validity in Theorem~\ref{thm:valid_p} calibrates the reference to $\mathcal{A}$ rather than to a finer partition by $\mathcal{S}(\bm{X})$.
At the level of sets, this aggregation is the union over the internal-state fibers compatible with $\mathcal{A}$, as expressed in Equation~\eqref{eq:Z_union_marginalization}; it is not an additional probabilistic conditioning event.
When aggregation implements the union truncation $\mathcal{Z}$, replacing the single local interval $I_{z_{\mathrm{obs}}}$ with $\mathcal{Z}$ preserves the validity of the conditional pivot.
The operational Algorithm~\ref{alg:parametric-si} instantiates this stitching over the connected components of the piecewise-quadratic fixed-state sets.

\newpage
\section{Identifying the conditional interval for $k$-means cluster assignments}
\label{appendix:kmeans_interval}
\label{appendix:kmeans_constraints_z}

This appendix details the construction of the local interval $[L_{\mathrm{KM}}, U_{\mathrm{KM}}]$ containing a current search point on which the $k$-means assignment path is unchanged, complementing the local-interval characterization in Section~\ref{sec:method} (Equation~\eqref{eq:local_interval_intersection}).

\subsection{Linear representation of the latent features and centroid dependence on the line coordinate}

Fix a current search point $z_0$ in Algorithm~\ref{alg:parametric-si}, and use $r$ for a candidate coordinate on the search line.
After fixing the nuisance sufficient statistic, the candidate data are
\begin{equation}
\bm{X}(r) = \bm{a} + \bm{b} r,
\label{eq:line_search}
\end{equation}
so that sample $i$ moves along $\bm{X}_i(r) = \bm{a}_i + \bm{b}_i r$, where $\bm{a}_i, \bm{b}_i \in \mathbb{R}^D$ are the $i$-th blocks of $\bm{a}$ and $\bm{b}$.
The $k$-means step, however, operates not on these observations but on the latent representations produced by the encoder.
Let $\mathcal{R}_{\rho}$ be the activation region containing $\bm{X}(z_0)$.
As established in Problem~A, on the component $[L_{\mathrm{NN}}, U_{\mathrm{NN}}]$ containing $z_0$, the encoder reduces to the affine map $(\bm{W}^{(\rho)}, \bm{c}^{(\rho)})$ of \eqref{eq:affine_map_general}, so the latent representations are affine in $r$:
\begin{equation}
\bm{\zeta}(r)
=
f_{\phi}(\bm{X}(r))
=
\bm{W}^{(\rho)} (\bm{a} + \bm{b} r) + \bm{c}^{(\rho)}
=
\tilde{\bm{a}} + \tilde{\bm{b}} r,
\label{eq:latent_line_app}
\end{equation}
and we write $\bm{\zeta}_i(r) = \tilde{\bm{a}}_i + \tilde{\bm{b}}_i r \in \mathbb{R}^d$ for the $i$-th block.
The affine formula is extended algebraically when solving the $k$-means constraints, but it equals the actual encoder output only on $[L_{\mathrm{NN}}, U_{\mathrm{NN}}]$; this is why the solutions of Problems~A and~B are intersected in \eqref{eq:local_interval_intersection}.

Run Lloyd's algorithm at the anchor data $\bm{X}(z_0)$, and let $\mathcal{C}_k^{(t)}(z_0)$ be the index set assigned to cluster $k$ at iteration $t$.
When deriving the interval around $z_0$, these anchor assignments are held fixed.
The corresponding candidate centroid is affine in $r$:
\begin{equation}
\bm{m}_k^{(t)}(r;z_0)
=
\frac{1}{|\mathcal{C}_k^{(t)}(z_0)|}
\sum_{i \in \mathcal{C}_k^{(t)}(z_0)}
(\tilde{\bm{a}}_i + \tilde{\bm{b}}_i r)
=
\bar{\bm{a}}_{k,z_0}^{(t)}
+
\bar{\bm{b}}_{k,z_0}^{(t)} r,
\label{eq:centroid_affine_app}
\end{equation}
where $\bar{\bm{a}}_{k,z_0}^{(t)}$ and $\bar{\bm{b}}_{k,z_0}^{(t)}$ do not depend on the candidate coordinate $r$.
The iteration is initialized with a fixed random label assignment that does not depend on the line coordinate, so the initial centroids have the same affine form.
If the fixed empty-cluster rule selects sample $i_0$ during the anchor run, this choice is recorded in the symbolic path and the candidate centroid is $\bm{m}_k^{(t)}(r;z_0)=\bm{\zeta}_{i_0}(r)=\tilde{\bm{a}}_{i_0}+\tilde{\bm{b}}_{i_0}r$, which remains affine in $r$.

\subsection{Expanding distance inequalities into quadratic forms}

For the assignments at a candidate coordinate $r$ to match those in the anchor run at $z_0$, the squared distance from $\bm{\zeta}_i(r)$ to its anchor-assigned centroid must be no larger than the distance to any competing centroid at every iteration.
The assignment condition is
\begin{equation}
\| \bm{\zeta}_i(r) - \bm{m}_{y_i^{(t+1)}(z_0)}^{(t)}(r;z_0) \|^2
\le
\| \bm{\zeta}_i(r) - \bm{m}_k^{(t)}(r;z_0) \|^2,
\quad \forall k \ne y_i^{(t+1)}(z_0),
\label{eq:dist_inequality_app}
\end{equation}
where $y_i^{(t+1)}(z_0)$ is the label assigned to point $i$ at iteration $t{+}1$ of the run on $\bm{X}(z_0)$.

Define $q_{i,k}^{(t)}(r;z_0) = \| \bm{\zeta}_i(r) - \bm{m}_k^{(t)}(r;z_0) \|^2$ and set
$\Delta \bm{a}_{i,k}^{(t)}(z_0) = \tilde{\bm{a}}_i - \bar{\bm{a}}_{k,z_0}^{(t)}$ and
$\Delta \bm{b}_{i,k}^{(t)}(z_0) = \tilde{\bm{b}}_i - \bar{\bm{b}}_{k,z_0}^{(t)}$.
Then
\begin{align}
q_{i,k}^{(t)}(r;z_0)
&= \| \Delta \bm{a}_{i,k}^{(t)}(z_0) + \Delta \bm{b}_{i,k}^{(t)}(z_0) r \|^2 \nonumber \\
&= \bigl(\Delta \bm{b}_{i,k}^{(t)}(z_0)^\top \Delta \bm{b}_{i,k}^{(t)}(z_0)\bigr) r^2 \nonumber \\
&\quad
+ 2\bigl(\Delta \bm{a}_{i,k}^{(t)}(z_0)^\top \Delta \bm{b}_{i,k}^{(t)}(z_0)\bigr) r
+ \Delta \bm{a}_{i,k}^{(t)}(z_0)^\top \Delta \bm{a}_{i,k}^{(t)}(z_0).
\end{align}
Let $A_{i,k}^{(t)}(z_0)$, $B_{i,k}^{(t)}(z_0)$, and $C_{i,k}^{(t)}(z_0)$ denote the coefficients of $r^2$, $r$, and $1$ in $q_{i,k}^{(t)}(r;z_0)$, in the notation of \eqref{eq:squared_distance}.
Writing $\Delta A_{i,k}^{(t)}(z_0) = A_{i,y_i^{(t+1)}(z_0)}^{(t)}(z_0) - A_{i,k}^{(t)}(z_0)$ and analogously for $\Delta B_{i,k}^{(t)}(z_0)$ and $\Delta C_{i,k}^{(t)}(z_0)$, condition \eqref{eq:dist_inequality_app} is equivalent to
\begin{equation}
\Delta A_{i,k}^{(t)}(z_0) r^2
+ \Delta B_{i,k}^{(t)}(z_0) r
+ \Delta C_{i,k}^{(t)}(z_0)
\le 0.
\label{eq:kmeans_quadratic_final}
\end{equation}

\subsection{Combining invariance constraints over the full $k$-means run}

To fix the entire $k$-means execution path (internal state $\mathcal{S}_{\mathrm{KM}}$), all inequalities arising at every iteration must hold simultaneously.
Let $T(z_0)$ be the number of iterations in the anchor run and, for each $t \in \{0, \dots, T(z_0)-1\}$, let $\mathcal{Z}^{(t)}(z_0)$ be the set of candidate coordinates $r$ satisfying \eqref{eq:kmeans_quadratic_final} for all $i$ and all competing clusters $k$.
The feasible set for the complete anchor path is
\begin{equation}
\mathcal{Z}_{\mathrm{KM}}(z_0)
=
\bigcap_{t=0}^{T(z_0)-1}
\mathcal{Z}^{(t)}(z_0).
\end{equation}
In practice, one solves each quadratic inequality analytically on the line.
The solution set of \eqref{eq:kmeans_quadratic_final} is not necessarily an interval. Abbreviating its coefficients by $\Delta A$, $\Delta B$, and $\Delta C$, the solution is an interval or the empty set when $\Delta A>0$, the complement of an open interval or all of $\mathbb{R}$ when $\Delta A<0$, and a half-line when $\Delta A=0$ and $\Delta B\ne0$; if $\Delta A=\Delta B=0$, it is either all of $\mathbb{R}$ or the empty set. The applicable case is determined by the discriminant and the constant term.
The intersection over all $i$, $k$, and $t$ is therefore in general a finite union of intervals.
We define $[L_{\mathrm{KM}}(z_0),U_{\mathrm{KM}}(z_0)]$ as the connected component of $\mathcal{Z}_{\mathrm{KM}}(z_0)$ containing the anchor $z_0$.
This is the maximal range of the candidate coordinate $r$ around $z_0$ for which the quadratic assignment constraints reproduce the Lloyd path at $\bm{X}(z_0)$.
In Section~\ref{sec:method}, the dependence on $z_0$ is suppressed and this component is written as $[L_{\mathrm{KM}},U_{\mathrm{KM}}]$.

\newpage
\section{Additional experiments: robustness of Type~I error control}
\label{appendix:robustness}

Beyond the main synthetic study in Section~\ref{sec:ArtificialExperiment}, we ran further null experiments to check whether Type~I error control is preserved under departures from the ideal setup (independent Gaussian noise with known covariance).
The following subsections summarize plug-in variance estimation and non-Gaussian noise; in each case we focus on whether \texttt{Proposed} and related baselines remain near the nominal level~$\alpha$.

\subsection{Estimated variance}
\label{appendix:robust_estvar}

\begin{figure}[htbp]
    \centering
    \begin{subfigure}{0.45\textwidth}
        \centering
        \includegraphics[width=\textwidth]{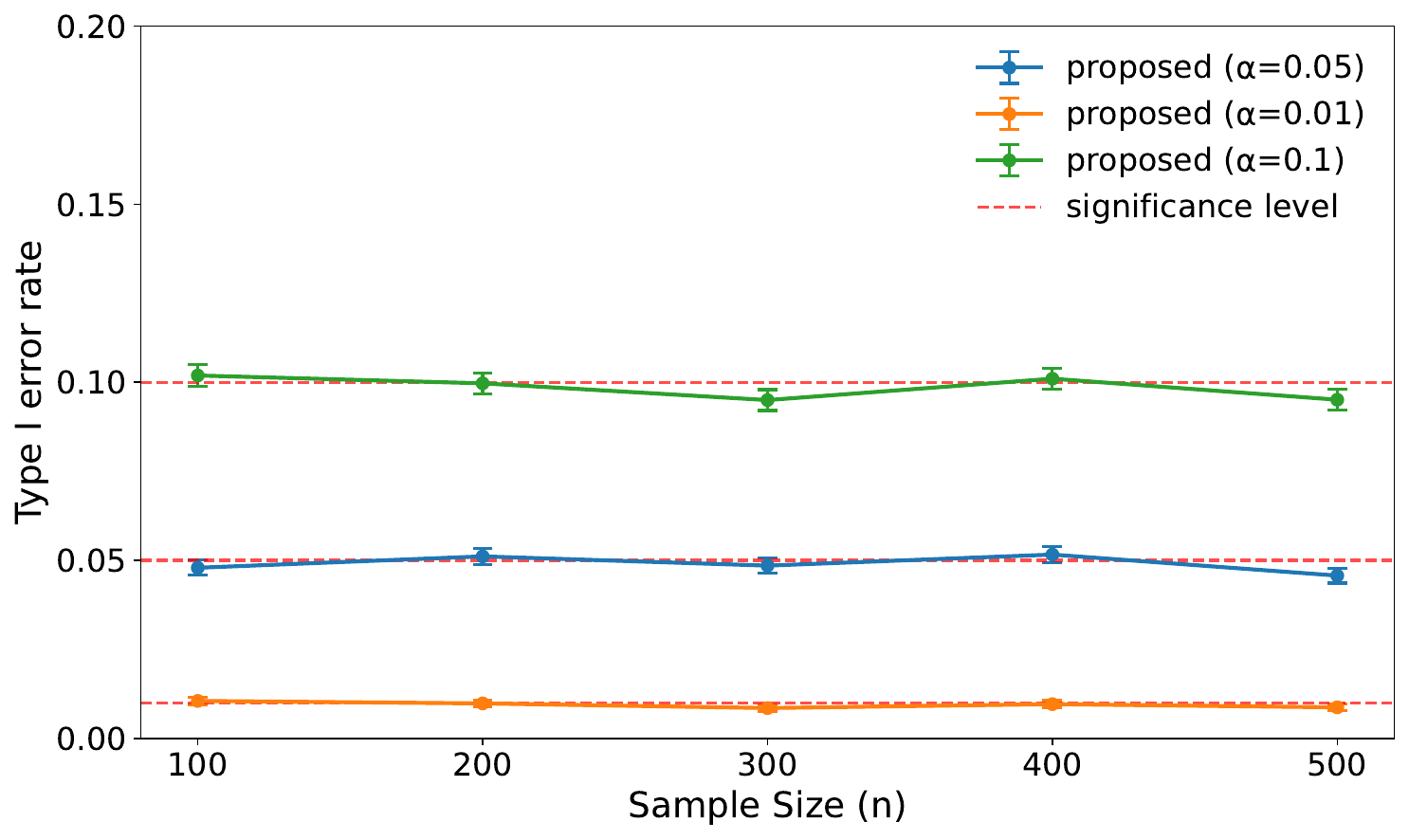}
        \caption{Type~I error rate with variance estimation (varying $N$)}
        \label{fig:fpr_n_robust}
    \end{subfigure}
    \hfill
    \begin{subfigure}{0.45\textwidth}
        \centering
        \includegraphics[width=\textwidth]{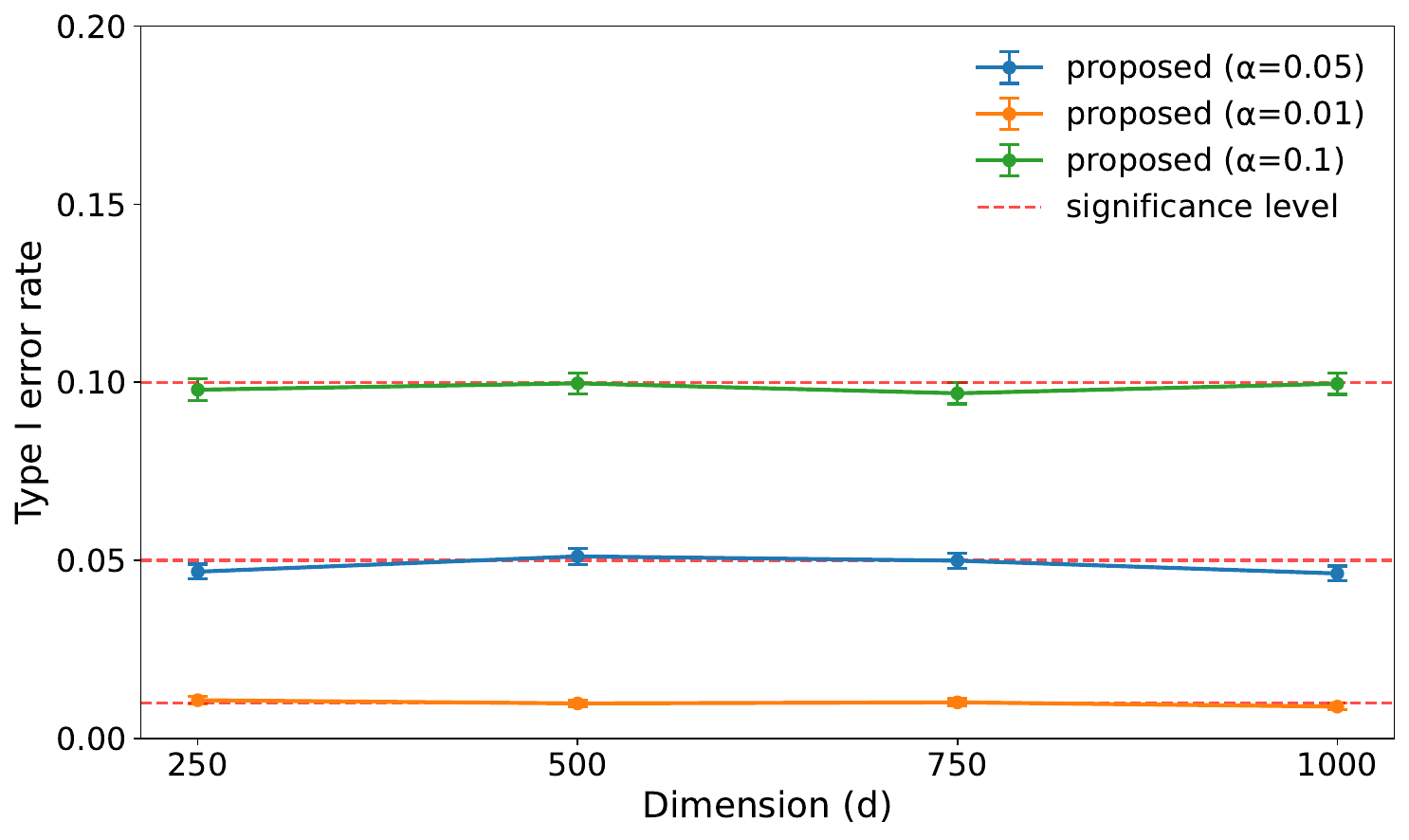}
        \caption{Type~I error rate with variance estimation (varying $D$)}
        \label{fig:fpr_d_robust}
    \end{subfigure}
    \caption{Robustness of Type~I error control when the variance is estimated from the data.}
    \label{fig:robust_estvar}
\end{figure}

We next assess Type~I error control when the noise variance is unknown and must be estimated from the data, a setting closer to practice than assuming a known $\bm{\Sigma}$.
Figure~\ref{fig:robust_estvar} summarizes the empirical Type~I error rate as $N$ and $D$ vary.
The empirical Type~I error rate remains near the nominal level in the tested settings when plug-in variance estimates are used, providing empirical evidence of robustness but not an exact guarantee for estimated covariance.

\subsection{Non-Gaussian noise}
\label{appendix:robust_nongauss}

\begin{figure}[htbp]
    \centering
    \includegraphics[width=0.45\textwidth]{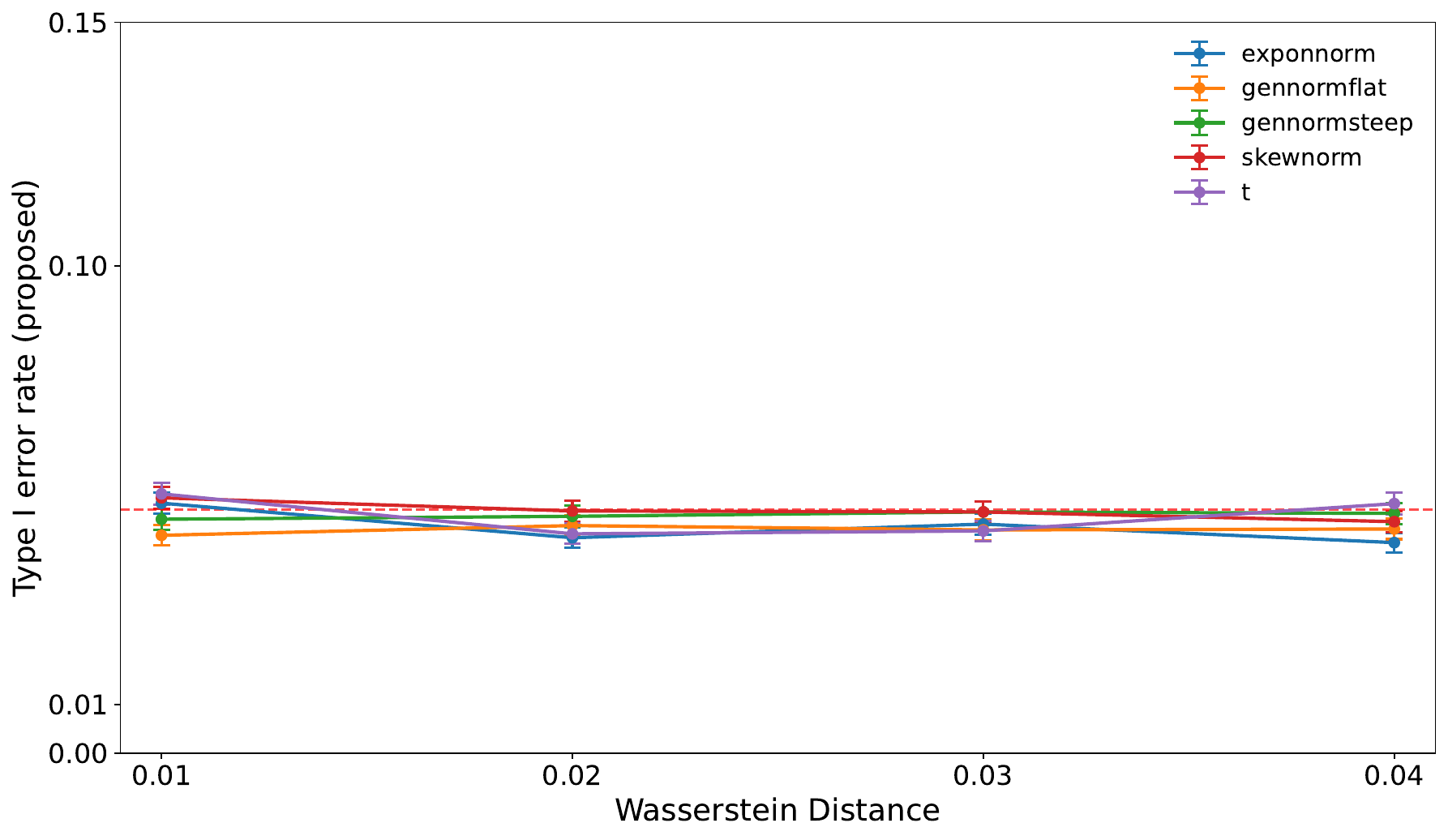}
    \caption{Type~I error rate under non-Gaussian noise families.}
    \label{fig:robust_nongauss}
\end{figure}

Selective inference for this model is derived under Gaussian errors.
To probe distributional misspecification, we generate null data from non-Gaussian families while keeping the remainder of the method unchanged.
Figure~\ref{fig:robust_nongauss} shows the empirical Type~I error rate.
Occasional excursions above $\alpha$ occur because the theory assumes normality; the results indicate empirical robustness in several tested conditions, but they do not extend the selective-validity guarantee beyond Gaussian errors.

\newpage
\section{Cluster center generation for power experiments}
\label{appendix:DataGeneration}

In the power experiments of Section~\ref{sec:ArtificialExperiment}, we construct cluster centers in high-dimensional space while preserving pairwise distances. The procedure is as follows.

\begin{enumerate}
    \item \textbf{Placement in a two-dimensional plane}

    First, on $\mathbb{R}^2$, we define vertices of an equilateral triangle with side length $\mathrm{sep}$ as rows of $\mathrm{base} \in \mathbb{R}^{3 \times 2}$:
    \begin{equation}
    \mathrm{base} = \begin{pmatrix} 0 & 0 \\ \mathrm{sep} & 0 \\ \mathrm{sep}/2 & \frac{\sqrt{3}}{2}\mathrm{sep} \end{pmatrix}.
    \end{equation}

    \item \textbf{Construction of an orthonormal basis}

    To map the two-dimensional structure into a subspace of $\mathbb{R}^D$, we generate a matrix $G \in \mathbb{R}^{D \times 2}$ with independent standard normal entries and apply the QR decomposition, letting $\Theta \in \mathbb{R}^{D \times D}$ denote its orthogonal factor. We take the first two columns $V = \Theta_{:, 1:2} \in \mathbb{R}^{D \times 2}$, which form an orthonormal basis:
    \begin{equation}
    V^\top V = I_2.
    \end{equation}

    \item \textbf{Distance preservation}

    For vertices $\bm{p}_i, \bm{p}_j \in \mathbb{R}^2$ (rows of $\mathrm{base}$), we embed them as $\tilde{\bm{p}}_i = \bm{p}_i V^\top \in \mathbb{R}^D$. Then
    \begin{align}
    \|\tilde{\bm{p}}_i - \tilde{\bm{p}}_j\|^2 &= \|(\bm{p}_i - \bm{p}_j) V^\top\|^2 \notag \\
    &= (\bm{p}_i - \bm{p}_j) V^\top V (\bm{p}_i - \bm{p}_j)^\top \notag \\
    &= \|\bm{p}_i - \bm{p}_j\|^2 = \mathrm{sep}^2.
    \end{align}
    Thus the separation $\mathrm{sep}$ between cluster centers is preserved after embedding in $\mathbb{R}^D$.
\end{enumerate}

\newpage
\section{Supplementary biological discussion for real-data experiments}
\label{appendix:realdata_biology}

This appendix collects extended biological context and gene-level interpretation for the real-data experiments in Section~\ref{subsec:real}.

\subsection{SEQC (neuroblastoma): context and interpretation}
\label{appendix:seqc_biology}

The SEQC neuroblastoma dataset consists of bulk RNA-seq profiles of neuroblastoma samples (GEO: GSE49711) \citep{zhang2015comparison}. Because expression is measured at the tissue level rather than at single-cell resolution, multiple factors such as tumor purity and admixture of normal neural tissue can jointly affect the signal \citep{yoshihara2013inferring, shimada1999international}. We therefore do not interpret clusters as definitive biological subtypes, but use them to summarize major data-driven variation.

\begin{figure}[htbp]
    \centering
    \begin{minipage}[b]{0.48\textwidth}
        \centering
        \includegraphics[width=\textwidth]{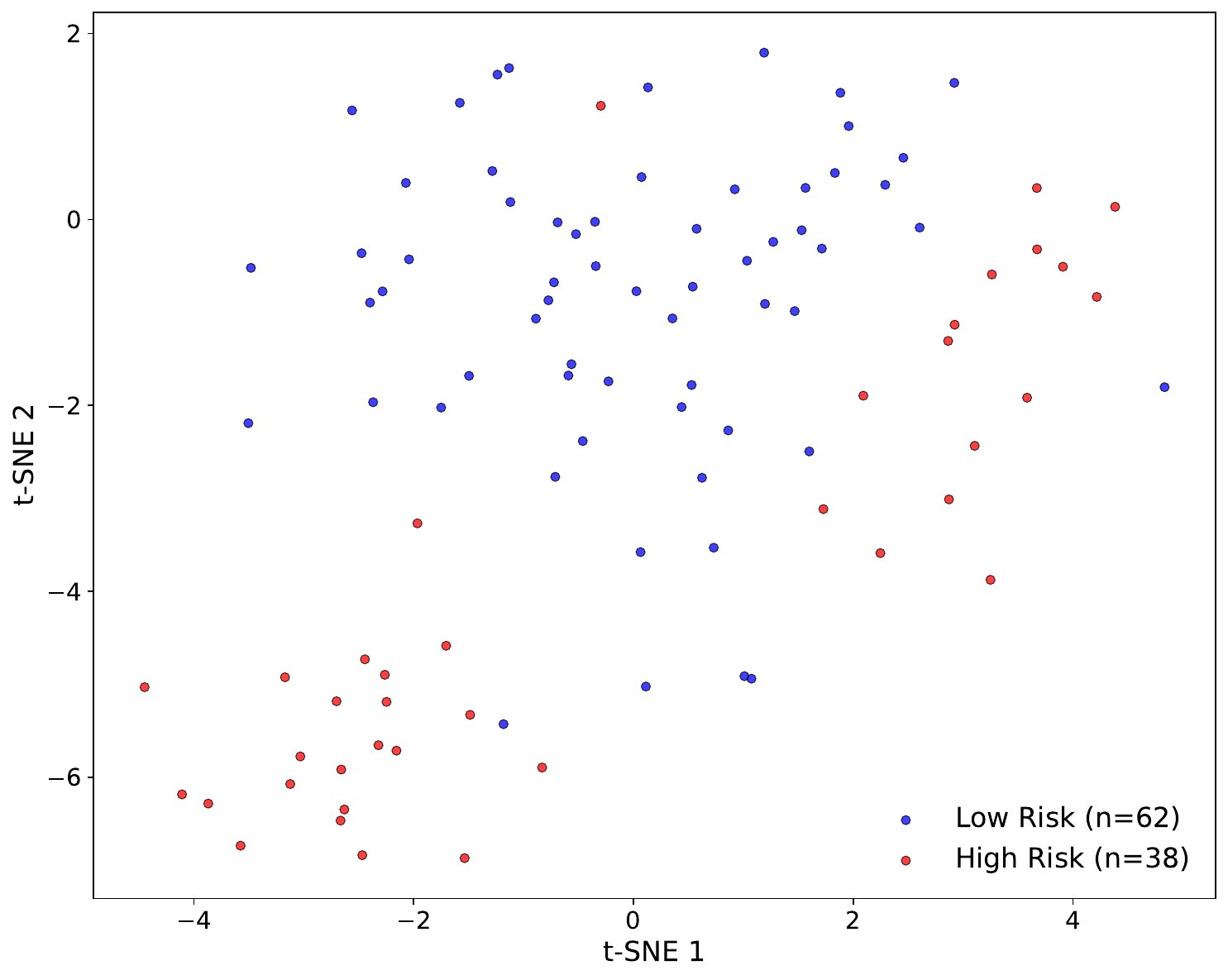}
        \subcaption{Ground Truth}
    \end{minipage}
    \hfill
    \begin{minipage}[b]{0.48\textwidth}
        \centering
        \includegraphics[width=\textwidth]{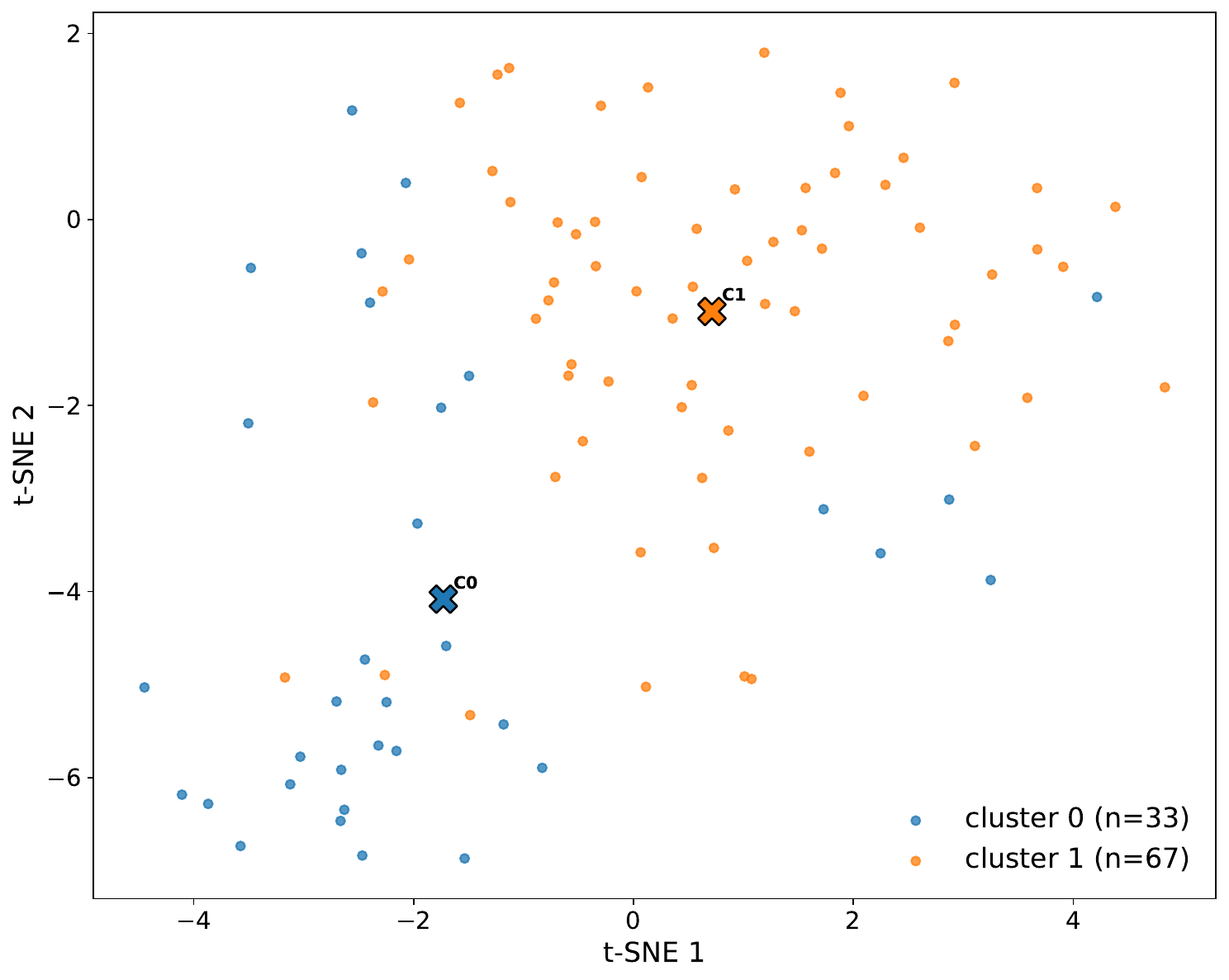}
        \subcaption{$K=2$}
    \end{minipage}
    \hfill
    \begin{minipage}[b]{0.48\textwidth}
        \centering
        \includegraphics[width=\textwidth]{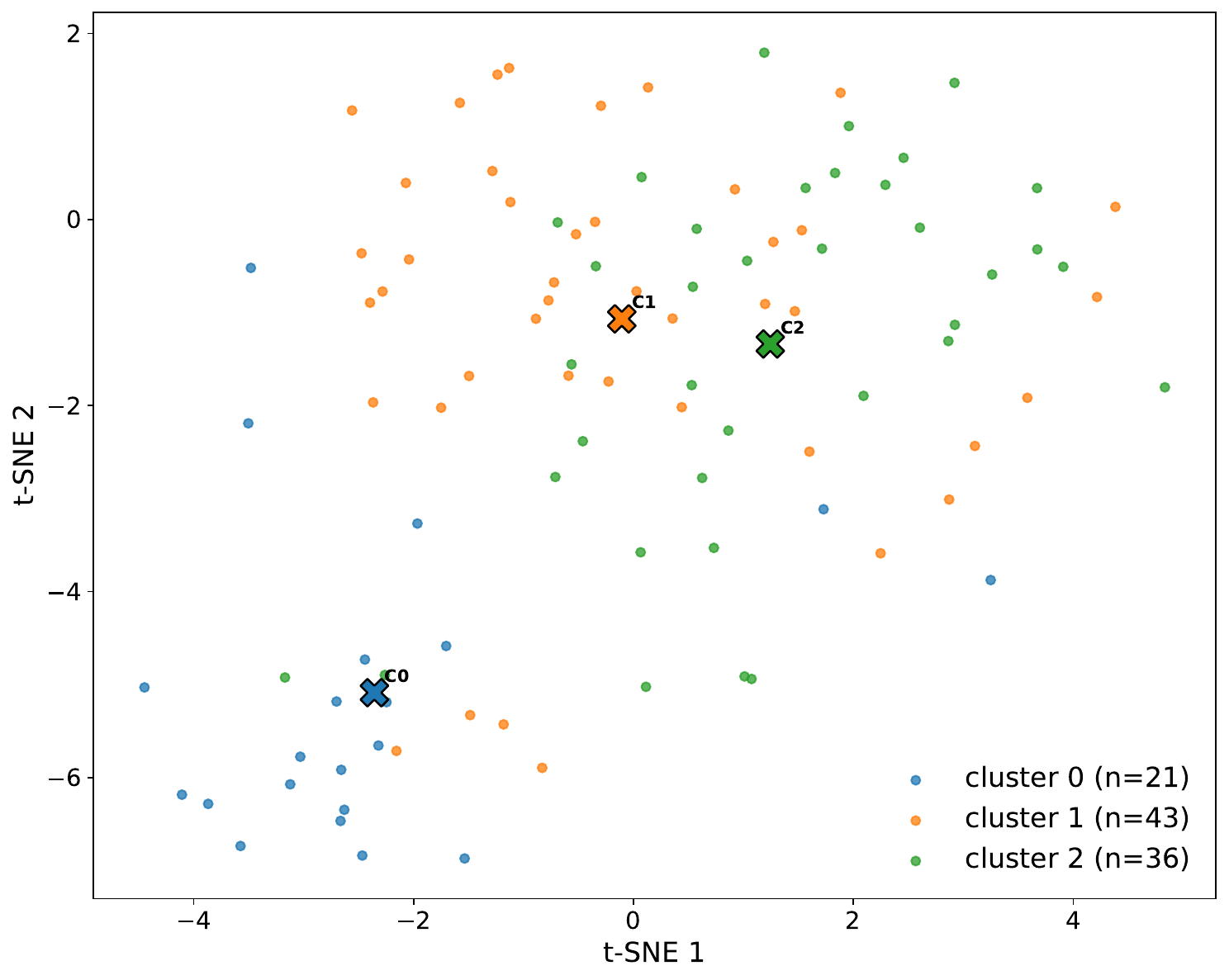}
        \subcaption{$K=3$}
    \end{minipage}
    \hfill
    \begin{minipage}[b]{0.48\textwidth}
        \centering
        \includegraphics[width=\textwidth]{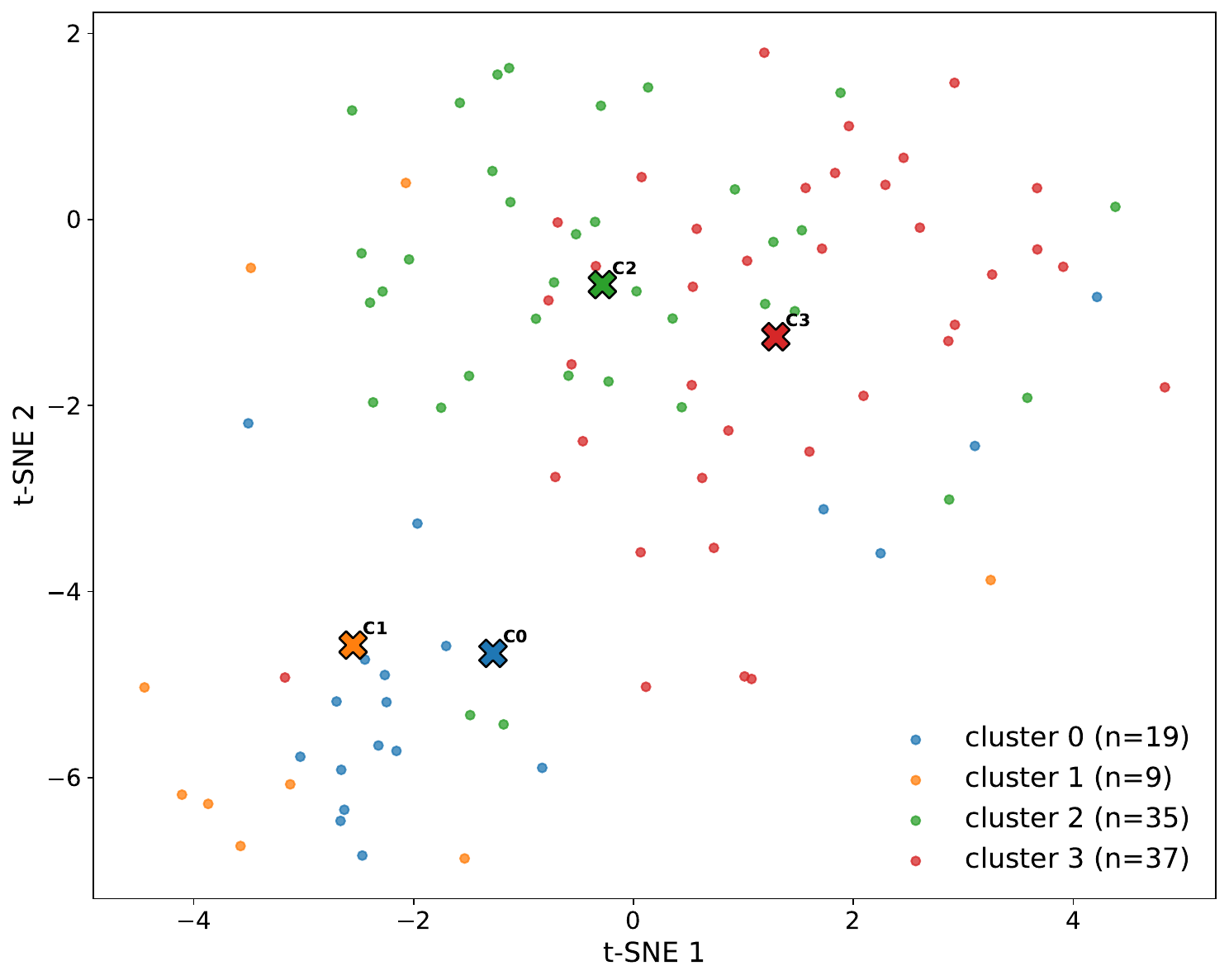}
        \subcaption{$K=4$}
    \end{minipage}

    \caption{t-SNE visualization of the input data space (100 dimensions). (a) Color-coded by clinical risk labels (Red: High-risk, Blue: Low-risk). (b)--(d) Color-coded based on deep clustering results.}
    \label{fig:tsne_visualization}
    \label{fig:tsne_appendix}
\end{figure}

Figure~\ref{fig:tsne_visualization} compresses the 100-dimensional input with t-SNE. Comparing risk labels in panel (a) with the $K=2$ clustering in (b) shows that the partition roughly tracks the main separation in the embedded space. For $K=3$ (c) and $K=4$ (d), the region occupied by the low-risk group in (a) is further split. t-SNE emphasizes local distances, so visual separation does not prove distinct biological subtypes. The subdivisions inside the low-risk group may reflect intrinsic tumor heterogeneity, specimen purity, or non-tumor contamination; the data alone do not identify the cause. At minimum, they show that latent expression structure not fully explained by the clinical risk label exists even within the low-risk stratum, and that deep clustering can pick up such fine-grained variation.

\subsubsection{Risk-group composition by cluster (SEQC)}
\label{appendix:seqc_cluster_composition}
\label{appendix:cluster_details}

Table~\ref{tab:cluster_composition} reports the composition ratio of clinical risk groups within each cluster obtained in the Risk-mixed setting. In the $K=2$ case, for example, one cluster is high-risk-dominant (approximately 73\%) while the other is low-risk-dominant (approximately 79\%), indicating that the learned partition reflects major risk-related variation to some extent.

\begin{table}[htbp]
    \centering
    \caption{Risk group composition ratio of each cluster in the Risk-mixed setting}
    \label{tab:cluster_composition}
    \begin{tabular}{ccccc}
        \hline
        $K$ & Cluster & $N$ & Low-risk  & High-risk \\
        \hline
        2 & 0 & 33 & 9 (27.3\%) & \textbf{24 (72.7\%)} \\
         & 1 & 67 & \textbf{53 (79.1\%)} & 14 (20.9\%) \\
        \hline
        3 & 0 & 21 & 2 (9.5\%) & \textbf{19 (90.5\%)} \\
        & 1 & 43 & \textbf{33 (76.7\%)} & 10 (23.3\%) \\
          & 2 & 36 & \textbf{27 (75.0\%)} & 9 (25.0\%) \\
        \hline
        4 & 0 & 19 & 1 (5.3\%) & \textbf{18 (94.7\%)} \\
         & 1 & 9 & 2 (22.2\%) & \textbf{7 (77.8\%)} \\
          & 2 & 35 & \textbf{30 (85.7\%)} & 5 (14.3\%) \\
          & 3 & 37 & \textbf{29 (78.4\%)} & 8 (21.6\%) \\
        \hline
    \end{tabular}
    \vspace{1mm} \\
    {\footnotesize \textbf{Bold} text indicates the dominant risk group in each cluster.}
\end{table}

\subsubsection{Gene-level evaluation (SEQC)}

We examine individual genes to assess face validity of the testing outcomes. Tables~\ref{tab:target_genes_pvalues_data5} and \ref{tab:target_genes_pvalues_data4} report $p$-values for representative genes.

As an overall trend, the \texttt{naive} method flags many low-expression genes with poorly annotated functions (e.g., LOC100288273). In bulk RNA-seq, low-expression genes are easily perturbed by technical noise, so many of these hits are plausibly non-biological \citep{law2014voom}.

In the Risk-mixed setting (Table~\ref{tab:target_genes_pvalues_data4}), the proposed method marks NTRK1 (TrkA) \citep{brodeur2003neuroblastoma}, a marker linked to favorable differentiation and prognosis in neuroblastoma, as significant. Purity and normal-tissue effects cannot be excluded in bulk data; still, recovering a cluster pattern consistent with the known low-risk/TrkA-high profile supports interpretability. \texttt{w/o-pp} often misses this gene, indicating limited power. The proposed method also finds differences for DLL3, ERBB3, and CAMK2A, which are involved in neural development and may reflect maturity differences between risk groups or varying normal-tissue content \citep{janoueix2010gene}. Whether the driver is intrinsic tumor biology or microenvironment, separating such pathology-related shifts from noise is consistent with useful selective inference.

In the Low-risk-only setting (Table~\ref{tab:target_genes_pvalues_data5}), TARP (TCR $\gamma$ alternate reading frame protein) illustrates method behavior. TARP is TCR-related; significant signals in Risk-mixed data align with reports that immune context differs by risk \citep{mina2015tumor, wienke2021integrative}. Within Low-risk-only, \texttt{naive} often calls TARP significant whereas the proposed method usually does not. Immune infiltration and stromal fractions vary across low-risk samples, so \texttt{naive} hits need not be ``wrong,'' yet strong prognostic splits as in Risk-mixed are less plausible within a single clinical low-risk stratum; aggressive \texttt{naive} calling may over-interpret small individual differences. The proposed method's conservatism here is consistent with guarding against differences that do not map cleanly to major clinical strata. Under some conditions the proposed method detects FOSB, KLF4, ERBB3, and NR4A3; at $\alpha = 0.05$ some may be Type~I errors, but these loci are also tied to differentiation, growth control, and stress responses, so mild within--low-risk biology remains conceivable. Relative to \texttt{naive}, the proposed method targets a narrower, more interpretable set of genes and reduces indiscriminate positives.

\begingroup
\footnotesize
\begin{longtable}{lccccc}
    \caption{$p$-values of representative genes in the Low-risk-only setting} \label{tab:target_genes_pvalues_data5} \\
    \hline
    Gene & $K$ & Cluster Pair & naive & w/o-pp & proposed \\
    \hline
    \endfirsthead

    \multicolumn{6}{c}{Continuation of {\tablename} \thetable{}} \\
    \hline
    Gene & $K$ & Cluster Pair & naive & w/o-pp & proposed \\
    \hline
    \endhead
    \hline
    \multicolumn{6}{r}{{Continued on next page}} \\
    \endfoot
    \hline
    \endlastfoot
    LOC100288273 & 3 & 0-1 & \textcolor{red}{<.001} &.772 &.978 \\
    LOC100288273 & 3 & 0-2 & \textcolor{red}{<.001} &.338 &.324 \\
    LOC100288273 & 4 & 0-1 & \textcolor{red}{.009} &.314 &.314 \\
    LOC100288273 & 4 & 0-2 & \textcolor{red}{.005} &.535 &.580 \\
    LOC100288273 & 4 & 0-3 & \textcolor{red}{.009} &.468 &.327 \\
    LOC100288273 & 4 & 1-2 & \textcolor{red}{<.001} &.151 &.623 \\
    \hline
    TARP & 2 & 0-1 & \textcolor{red}{<.001} &.157 &.170 \\
    TARP & 3 & 0-1 & \textcolor{red}{<.001} &.759 &.746 \\
    TARP & 3 & 0-2 & \textcolor{red}{.003} &.672 &.368 \\
    TARP & 4 & 0-1 & \textcolor{red}{.033} &.584 &.584 \\
    TARP & 4 & 0-2 & \textcolor{red}{.037} &.328 &.901 \\
    TARP & 4 & 1-2 & \textcolor{red}{<.001} &.399 &.435 \\
    \hline
    FOSB & 2 & 0-1 & \textcolor{red}{.003} &.163 & \textcolor{red}{.003} \\
    FOSB & 3 & 0-1 & \textcolor{red}{<.001} &.920 &.910 \\
    FOSB & 3 & 1-2 & \textcolor{red}{<.001} &.976 &.555 \\
    FOSB & 4 & 0-1 & \textcolor{red}{.003} &.367 &.144 \\
    FOSB & 4 & 1-2 & \textcolor{red}{<.001} &.735 &.162 \\
    FOSB & 4 & 1-3 & \textcolor{red}{.012} &.557 &.528 \\
    KLF4 & 2 & 0-1 & \textcolor{red}{<.001} &.550 &.556 \\
    KLF4 & 3 & 0-1 & \textcolor{red}{<.001} &.989 & \textcolor{red}{.001} \\
    KLF4 & 3 & 1-2 & \textcolor{red}{.005} &.937 &.050 \\
    KLF4 & 4 & 0-1 & \textcolor{red}{.001} &.308 &.601 \\
    KLF4 & 4 & 1-2 & \textcolor{red}{<.001} &.207 & \textcolor{red}{.015} \\
    KLF4 & 4 & 1-3 & \textcolor{red}{.028} &.987 &.997 \\
    ERBB3 & 2 & 0-1 & \textcolor{red}{<.001} &.210 & \textcolor{red}{.014} \\
    ERBB3 & 3 & 0-1 & \textcolor{red}{<.001} &.288 &.614 \\
    ERBB3 & 4 & 1-2 & \textcolor{red}{<.001} &.405 &.275 \\
    NR4A3 & 2 & 0-1 & \textcolor{red}{.001} & \textcolor{red}{.004} & \textcolor{red}{.001} \\
    NR4A3 & 3 & 0-1 & \textcolor{red}{<.001} &.577 &.577 \\
    NR4A3 & 3 & 1-2 & \textcolor{red}{<.001} &.522 &.536 \\
    NR4A3 & 4 & 0-1 & \textcolor{red}{.002} &.142 &.249 \\
    NR4A3 & 4 & 1-2 & \textcolor{red}{<.001} &.806 & \textcolor{red}{.039} \\
    NR4A3 & 4 & 1-3 & \textcolor{red}{.020} &.352 &.291 \\
\end{longtable}
\endgroup

\begingroup
\footnotesize
\begin{longtable}{lccccc}
    \caption{$p$-values of representative genes in the Risk-mixed setting}
    \label{tab:target_genes_pvalues_data4} \\
    \hline
    Gene & $K$ & Cluster Pair & naive & w/o-pp & proposed \\
    \hline
    \endfirsthead
    \multicolumn{6}{c}{Continuation of {\tablename} \thetable{}} \\
    \hline
    Gene & $K$ & Cluster Pair & naive & w/o-pp & proposed \\
    \hline
    \endhead
    \hline
    \multicolumn{6}{r}{{Continued on next page}} \\
    \endfoot
    \hline
    \endlastfoot
    NTRK1 & 2 & 0-1 & \textcolor{red}{<.001} & \textcolor{red}{.005} & \textcolor{red}{<.001} \\
    NTRK1 & 3 & 0-1 & \textcolor{red}{<.001} &.246 & \textcolor{red}{.001} \\
    NTRK1 & 3 & 0-2 & \textcolor{red}{<.001} &.857 & \textcolor{red}{<.001} \\
    NTRK1 & 4 & 0-2 & \textcolor{red}{<.001} & \textcolor{red}{.033} & \textcolor{red}{<.001} \\
    NTRK1 & 4 & 0-3 & \textcolor{red}{<.001} & \textcolor{red}{.008} & \textcolor{red}{.017} \\
    NTRK1 & 4 & 1-2 & \textcolor{red}{<.001} &.977 & \textcolor{red}{.008} \\
    NTRK1 & 4 & 1-3 & \textcolor{red}{.002} &.981 & \textcolor{red}{.011} \\
    \hline
    DLL3 & 2 & 0-1 & \textcolor{red}{.003} &.482 & \textcolor{red}{.004} \\
    DLL3 & 3 & 0-1 & \textcolor{red}{.001} &.710 & \textcolor{red}{.003} \\
    DLL3 & 3 & 0-2 & \textcolor{red}{.002} &.610 & \textcolor{red}{.001} \\
    DLL3 & 4 & 0-2 & \textcolor{red}{<.001} & \textcolor{red}{.003} & \textcolor{red}{.008} \\
    DLL3 & 4 & 0-3 & \textcolor{red}{.001} & \textcolor{red}{.009} & \textcolor{red}{.029} \\
    ERBB3 & 2 & 0-1 & \textcolor{red}{<.001} & \textcolor{red}{.004} & \textcolor{red}{.001} \\
    ERBB3 & 3 & 0-1 & \textcolor{red}{<.001} &.662 & \textcolor{red}{<.001} \\
    ERBB3 & 3 & 0-2 & \textcolor{red}{<.001} &.257 & \textcolor{red}{.006} \\
    ERBB3 & 4 & 0-2 & \textcolor{red}{<.001} &.680 & \textcolor{red}{.012} \\
    ERBB3 & 4 & 0-3 & \textcolor{red}{<.001} & \textcolor{red}{.032} & \textcolor{red}{.001} \\
    ERBB3 & 4 & 1-2 & \textcolor{red}{.016} &.055 &.071 \\
    ERBB3 & 4 & 1-3 & \textcolor{red}{.002} & \textcolor{red}{.046} & \textcolor{red}{.011} \\
    CAMK2A & 2 & 0-1 & \textcolor{red}{<.001} & \textcolor{red}{.004} & \textcolor{red}{<.001} \\
    CAMK2A & 3 & 0-1 & \textcolor{red}{<.001} &.113 & \textcolor{red}{<.001} \\
    CAMK2A & 3 & 0-2 & \textcolor{red}{<.001} &.918 & \textcolor{red}{.002} \\
    CAMK2A & 4 & 0-2 & \textcolor{red}{<.001} &.410 & \textcolor{red}{.002} \\
    CAMK2A & 4 & 0-3 & \textcolor{red}{<.001} &.202 & \textcolor{red}{.002} \\
    CAMK2A & 4 & 1-2 & \textcolor{red}{<.001} &.619 & \textcolor{red}{.005} \\
    CAMK2A & 4 & 1-3 & \textcolor{red}{<.001} &.501 & \textcolor{red}{.005} \\
    \hline
    SEMA3D & 3 & 0-1 & \textcolor{red}{.011} &.662 & \textcolor{red}{.031} \\
    SEMA3D & 3 & 0-2 & \textcolor{red}{.016} &.842 &.138 \\
    SEMA3D & 4 & 1-2 & \textcolor{red}{.011} &.327 &.375 \\
    SEMA3D & 4 & 1-3 & \textcolor{red}{.029} &.158 &.387 \\
    TARP & 2 & 0-1 & \textcolor{red}{<.001} &.057 & \textcolor{red}{<.001} \\
    TARP & 3 & 0-1 & \textcolor{red}{<.001} &.804 & \textcolor{red}{.008} \\
    TARP & 3 & 0-2 & \textcolor{red}{<.001} &.920 &.087 \\
    TARP & 4 & 0-1 & \textcolor{red}{.002} &.051 & \textcolor{red}{.024} \\
    TARP & 4 & 0-2 & \textcolor{red}{.024} &.394 &.624 \\
    TARP & 4 & 0-3 & \textcolor{red}{.001} &.278 &.407 \\
    TARP & 4 & 1-2 & \textcolor{red}{<.001} & \textcolor{red}{.001} & \textcolor{red}{<.001} \\
    TARP & 4 & 1-3 & \textcolor{red}{<.001} & \textcolor{red}{.004} & \textcolor{red}{.004} \\
    ANO9 & 2 & 0-1 & \textcolor{red}{<.001} & \textcolor{red}{<.001} & \textcolor{red}{<.001} \\
    ANO9 & 3 & 0-1 & \textcolor{red}{<.001} &.090 & \textcolor{red}{.001} \\
    ANO9 & 3 & 0-2 & \textcolor{red}{<.001} &.373 & \textcolor{red}{.003} \\
    ANO9 & 3 & 1-2 & \textcolor{red}{.012} &.172 &.534 \\
    ANO9 & 4 & 0-1 & \textcolor{red}{<.001} &.943 &.123 \\
    ANO9 & 4 & 0-2 & \textcolor{red}{.002} &.273 &.351 \\
    ANO9 & 4 & 0-3 & \textcolor{red}{<.001} & \textcolor{red}{.041} & \textcolor{red}{.026} \\
    ANO9 & 4 & 1-2 & \textcolor{red}{<.001} &.128 & \textcolor{red}{<.001} \\
    ANO9 & 4 & 1-3 & \textcolor{red}{<.001} &.152 & \textcolor{red}{.003} \\
    ANO9 & 4 & 2-3 & \textcolor{red}{.048} &.621 &.154 \\
    GRIN3A & 2 & 0-1 & \textcolor{red}{<.001} & \textcolor{red}{.037} & \textcolor{red}{<.001} \\
    GRIN3A & 3 & 0-1 & \textcolor{red}{<.001} &.260 & \textcolor{red}{.002} \\
    GRIN3A & 3 & 0-2 & \textcolor{red}{<.001} &.484 & \textcolor{red}{.014} \\
    GRIN3A & 4 & 0-2 & \textcolor{red}{<.001} & \textcolor{red}{.043} & \textcolor{red}{.026} \\
    GRIN3A & 4 & 0-3 & \textcolor{red}{<.001} & \textcolor{red}{.042} &.054 \\
    GRIN3A & 4 & 1-2 & \textcolor{red}{<.001} &.136 &.246 \\
    GRIN3A & 4 & 1-3 & \textcolor{red}{.003} &.139 &.359 \\
    GRIN3A & 4 & 2-3 & \textcolor{red}{.018} &.623 & \textcolor{red}{.043} \\
    \hline
    LOC100216479 & 4 & 0-1 & \textcolor{red}{.004} &.163 &.157 \\
    LOC100216479 & 4 & 2-3 & \textcolor{red}{.049} &.885 &.097 \\
    LOC100287820 & 3 & 1-2 & \textcolor{red}{<.001} &.771 &.190 \\
    LOC100288273 & 3 & 0-1 & \textcolor{red}{.025} &.432 &.590 \\
    LOC100288273 & 3 & 0-2 & \textcolor{red}{<.001} &.219 &.069 \\
    LOC100288765 & 4 & 0-1 & \textcolor{red}{.048} &.661 &.208 \\
    LOC100288765 & 4 & 1-2 & \textcolor{red}{.019} &.487 &.120 \\
    LOC100288765 & 4 & 1-3 & \textcolor{red}{.017} &.376 &.662 \\
    LOC100289085 & 3 & 1-2 & \textcolor{red}{.049} &.986 &.968 \\
    LOC100289085 & 4 & 2-3 & \textcolor{red}{.041} &.263 &.211 \\
    LOC100289383 & 3 & 0-2 & \textcolor{red}{.008} &.869 &.811 \\
    LOC100289383 & 4 & 0-1 & \textcolor{red}{.033} &.718 &.121 \\
    LOC100289383 & 4 & 1-2 & \textcolor{red}{.015} &.949 &.057 \\
\end{longtable}
\endgroup

\subsection{PBMC: gene-level biological interpretation}
\label{appendix:pbmc_biology}

\subsubsection{Multi-type clustering visualization and quantitative summary}

\begin{figure}[htbp]
    \centering
    \begin{minipage}[b]{0.48\textwidth}
        \centering
        \includegraphics[width=\textwidth]{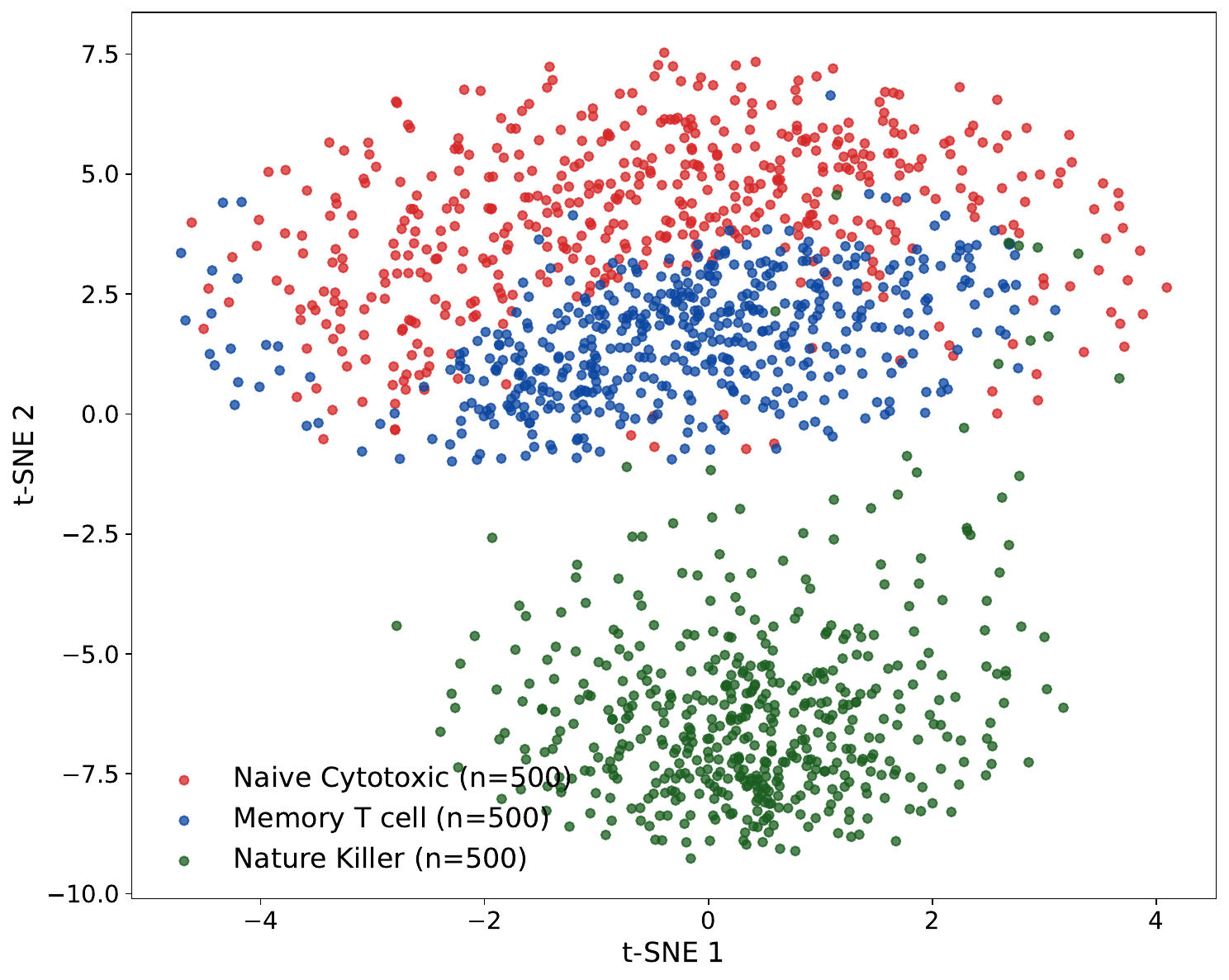}
        \subcaption{Ground Truth}
    \end{minipage}
    \hfill
    \begin{minipage}[b]{0.48\textwidth}
        \centering
        \includegraphics[width=\textwidth]{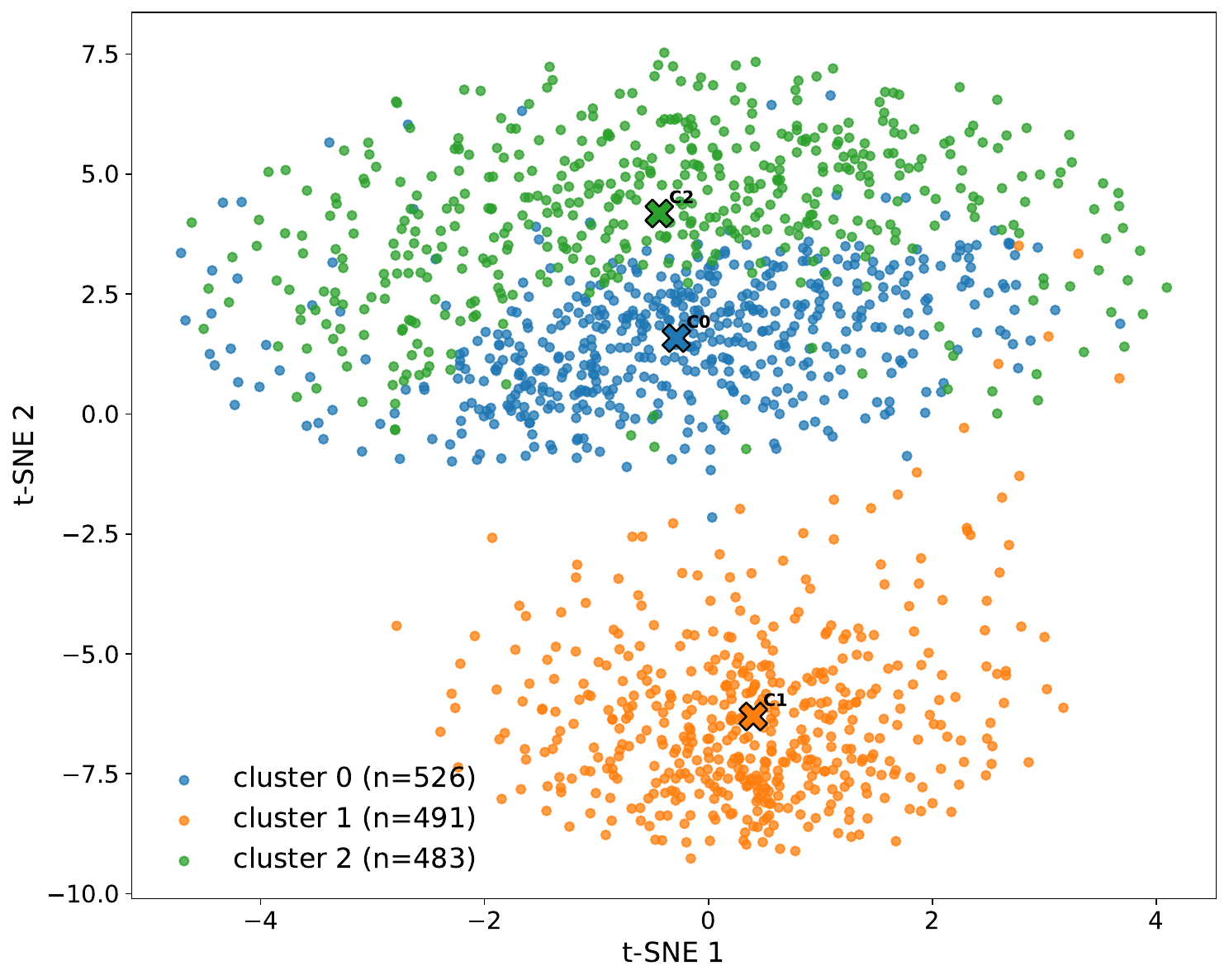}
        \subcaption{$K=3$ (Predicted)}
    \end{minipage}
    \caption{t-SNE visualization of the input data space (1000 dimensions) in the Multi-type setting. (a) True cell type labels, (b) Color-coding based on deep clustering ($K=3$).}
    \label{fig:tsne_visualization_multi}
\end{figure}

Figure~\ref{fig:tsne_visualization_multi} shows a two-dimensional t-SNE embedding of the 1000-dimensional inputs in the Multi-type setting.
Comparing the true cell-type labels (Fig.~\ref{fig:tsne_visualization_multi}a) with the deep clustering partition (Fig.~\ref{fig:tsne_visualization_multi}b) demonstrates that the learned clusters are highly consistent with established biological categories.
As a quantitative summary, we obtained an Adjusted Rand Index (ARI) of $0.899$.
Together, these results confirm that deep clustering successfully extracts discriminative features from high-dimensional expression profiles, yielding partitions that align closely with annotated lymphocyte types without the use of label supervision.

\subsubsection{Gene-level tables and interpretation}

Tables~\ref{tab:target_genes_pvalues_three_types} and \ref{tab:target_genes_pvalues_three_types_rinpa} list representative gene $p$-values in the Single-type and Multi-type settings. Genes that all three methods call non-significant ($p > .05$) are omitted.

In the Single-type setting (Memory T cells only), we expect few true cluster differences. Genes found only by \texttt{naive} include many low-expression RP11/RP4/RP5 transcripts; high variance and measurement noise make these likely false positives. In some pairs the proposed method detects ISG15 and RCC1 while \texttt{w/o-pp} does not. Memory T cells undergo homeostatic proliferation \citep{surh2008homeostasis}, so RCC1 (cell-cycle related) may track subtle division-rate differences. Interferon-response genes such as ISG15 also vary within immune populations \citep{shalek2013single}. Thus the proposed method suppresses dubious low-expression hits while retaining sensitivity to modest state changes (cycle or innate-response programs), though we cannot tell from this analysis alone whether those form stable clusters or transient states.

In the Multi-type setting, building on the cluster--label alignment summarized above, CDK11A is significant for all pairs under the proposed method but not under \texttt{w/o-pp} (Table~\ref{tab:target_genes_pvalues_three_types_rinpa}). Naive cells are metabolically quiet, whereas memory cells maintain higher baseline activation for rapid recall \citep{surh2008homeostasis}; the proposed test recovers this axis of functional difference that \texttt{w/o-pp} blunts.

For KIF2C, all methods yield very small $p$-values across pairs. KIF2C supports mitotic chromosome segregation and tracks proliferative activity. The three lineages differ in turnover (naive T lowest, memory T with slow self-renewal, NK relatively faster in periphery \citep{zhang2007vivo}), so broad significance matches expected proliferation contrasts.

ISG15 is strongest in pairs involving Cluster~1 (NK): 0--1 and 1--2. NK cells show higher basal interferon-program expression than adaptive T subsets \citep{vivier2011innate, kang2018multiplexed}, so Multi-type ISG15 patterns fit innate versus adaptive roles rather than the Single-type fluctuation story alone.

Overall, the proposed method limits selection-bias-driven noise seen under \texttt{naive}, while retaining sensitivity to differentiation- and function-related contrasts that \texttt{w/o-pp} often misses.

\begingroup
\footnotesize
\begin{longtable}{lccccc}
    \caption{$p$-values of representative genes in the Single-type setting}
    \label{tab:target_genes_pvalues_three_types} \\
    \hline
    Gene & $K$ & Cluster Pair & naive & w/o-pp & proposed \\
    \hline
    \endfirsthead

    \multicolumn{6}{c}{Continuation of {\tablename} \thetable{}} \\
    \hline
    Gene & $K$ & Cluster Pair & naive & w/o-pp & proposed \\
    \hline
    \endhead

    \hline
    \multicolumn{6}{r}{{Continued on next page}} \\
    \endfoot

    \hline
    \endlastfoot
    ISG15 & 3 & 0-1 & \textcolor{red}{.027} &.623 & \textcolor{red}{.050} \\
    ISG15 & 3 & 1-2 & \textcolor{red}{.006} &.508 &.728 \\
    RCC1 & 3 & 0-1 & \textcolor{red}{.026} &.480 & \textcolor{red}{.031} \\
    \hline
    RP11-108M9.1 & 3 & 0-1 & \textcolor{red}{<.001} &.485 &.476 \\
    RP11-108M9.1 & 3 & 0-2 & \textcolor{red}{.007} &.113 &.077 \\
    RP11-108M9.1 & 3 & 1-2 & \textcolor{red}{<.001} &.957 &.867 \\
    RP11-154H17.1 & 3 & 0-1 & \textcolor{red}{<.001} &.462 &.690 \\
    RP11-154H17.1 & 3 & 0-2 & \textcolor{red}{.026} &.072 &.199 \\
    RP11-154H17.1 & 3 & 1-2 & \textcolor{red}{<.001} &.087 &.125 \\
    RP11-174G17.2 & 3 & 0-1 & \textcolor{red}{.002} &.767 &.998 \\
    RP11-174G17.2 & 3 & 0-2 & \textcolor{red}{.006} &.848 &.093 \\
    RP11-174G17.2 & 3 & 1-2 & \textcolor{red}{<.001} &.721 &.243 \\
    RP11-181G12.2 & 3 & 0-1 & \textcolor{red}{<.001} &.615 &.585 \\
    RP11-181G12.2 & 3 & 0-2 & \textcolor{red}{.003} &.301 &.391 \\
    RP11-181G12.2 & 3 & 1-2 & \textcolor{red}{<.001} &.847 &.075 \\
    RP11-206L10.4 & 3 & 0-1 & \textcolor{red}{.046} &.317 &.701 \\
    RP11-206L10.4 & 3 & 0-2 & \textcolor{red}{.023} &.294 &.059 \\
    RP11-206L10.4 & 3 & 1-2 & \textcolor{red}{<.001} &.328 &.665 \\
    RP11-22L13.1 & 3 & 0-1 & \textcolor{red}{<.001} &.536 &.311 \\
    RP11-22L13.1 & 3 & 0-2 & \textcolor{red}{.015} &.719 &.882 \\
    RP11-22L13.1 & 3 & 1-2 & \textcolor{red}{<.001} &.925 &.327 \\
    RP11-293F5.1 & 3 & 0-1 & \textcolor{red}{.039} &.149 &.580 \\
    RP11-293F5.1 & 3 & 0-2 & \textcolor{red}{.010} &.186 &.255 \\
    RP11-293F5.1 & 3 & 1-2 & \textcolor{red}{<.001} &.506 &.268 \\
    RP11-345P4.7 & 3 & 0-1 & \textcolor{red}{.010} &.526 &.716 \\
    RP11-345P4.7 & 3 & 0-2 & \textcolor{red}{.015} &.192 &.360 \\
    RP11-345P4.7 & 3 & 1-2 & \textcolor{red}{<.001} &.892 &.219 \\
    RP11-420K8.1 & 3 & 0-1 & \textcolor{red}{<.001} &.391 &.816 \\
    RP11-420K8.1 & 3 & 0-2 & \textcolor{red}{.002} &.529 &.224 \\
    RP11-420K8.1 & 3 & 1-2 & \textcolor{red}{<.001} &.922 &.518 \\
    RP11-435D7.3 & 3 & 0-1 & \textcolor{red}{<.001} &.147 &.154 \\
    RP11-435D7.3 & 3 & 0-2 & \textcolor{red}{.013} &.456 &.532 \\
    RP11-435D7.3 & 3 & 1-2 & \textcolor{red}{<.001} &.754 &.104 \\
    RP11-547D24.1 & 3 & 0-1 & \textcolor{red}{.001} &.507 &.984 \\
    RP11-547D24.1 & 3 & 0-2 & \textcolor{red}{.001} &.663 &.616 \\
    RP11-547D24.1 & 3 & 1-2 & \textcolor{red}{<.001} &.855 &.258 \\
    RP11-54O7.17 & 3 & 0-1 & \textcolor{red}{<.001} &.780 &.861 \\
    RP11-54O7.17 & 3 & 0-2 & \textcolor{red}{<.001} &.575 &.187 \\
    RP11-54O7.17 & 3 & 1-2 & \textcolor{red}{<.001} &.917 &.249 \\
    \hline
\end{longtable}
\endgroup

\begingroup
\footnotesize
\begin{longtable}{lccccc}
    \caption{$p$-values of representative genes in the Multi-type setting}
    \label{tab:target_genes_pvalues_three_types_rinpa} \\
    \hline
    Gene & $K$ & Cluster Pair & naive & w/o-pp & proposed \\
    \hline
    \endfirsthead

    \multicolumn{6}{c}{Continuation of {\tablename} \thetable{}} \\
    \hline
    Gene & $K$ & Cluster Pair & naive & w/o-pp & proposed \\
    \hline
    \endhead

    \hline
    \multicolumn{6}{r}{{Continued on next page}} \\
    \endfoot

    \hline
    \endlastfoot
    CDK11A & 3 & 0-1 & \textcolor{red}{.031} &.300 & \textcolor{red}{.031} \\
    CDK11A & 3 & 0-2 & \textcolor{red}{<.001} &.469 & \textcolor{red}{<.001} \\
    CDK11A & 3 & 1-2 & \textcolor{red}{.004} &.572 & \textcolor{red}{.004} \\
    ISG15 & 3 & 0-1 & \textcolor{red}{.002} & \textcolor{red}{.050} & \textcolor{red}{.002} \\
    ISG15 & 3 & 1-2 & \textcolor{red}{<.001} & \textcolor{red}{<.001} & \textcolor{red}{<.001} \\
    KIF2C & 3 & 0-1 & \textcolor{red}{<.001} & \textcolor{red}{<.001} & \textcolor{red}{<.001} \\
    KIF2C & 3 & 0-2 & \textcolor{red}{<.001} & \textcolor{red}{.007} & \textcolor{red}{.007} \\
    KIF2C & 3 & 1-2 & \textcolor{red}{<.001} & \textcolor{red}{<.001} & \textcolor{red}{<.001} \\
    \hline
    RP11-34P13.14 & 3 & 0-2 & \textcolor{red}{.043} &.721 &.721 \\
    RP11-34P13.14 & 3 & 1-2 & \textcolor{red}{<.001} &.059 &.059 \\
    RP11-54O7.2 & 3 & 0-2 & \textcolor{red}{<.001} &.481 &.948 \\
    RP11-54O7.2 & 3 & 1-2 & \textcolor{red}{<.001} &.056 &.085 \\
    RP11-558F24.4 & 3 & 0-1 & \textcolor{red}{.019} &.058 &.079 \\
    RP11-567C20.2 & 3 & 0-2 & \textcolor{red}{<.001} &.148 &.214 \\
    RP11-567C20.3 & 3 & 0-2 & \textcolor{red}{.028} &.715 &.715 \\
    RP11-84A14.5 & 3 & 0-2 & \textcolor{red}{<.001} &.528 &.574 \\
    RP11-96L14.7 & 3 & 0-1 & \textcolor{red}{<.001} &.056 &.057 \\
    RP11-96L14.7 & 3 & 0-2 & \textcolor{red}{<.001} &.124 &.166 \\
    RP13-392I16.1 & 3 & 0-2 & \textcolor{red}{<.001} &.077 &.125 \\
    RP3-467K16.2 & 3 & 1-2 & \textcolor{red}{.011} &.058 &.073 \\
    RP3-467K16.4 & 3 & 0-2 & \textcolor{red}{<.001} &.067 &.067 \\
    RP3-510D11.2 & 3 & 0-2 & \textcolor{red}{<.001} &.241 &.241 \\
    RP4-533D7.4 & 3 & 0-2 & \textcolor{red}{.019} &.098 &.176 \\
    RP4-533D7.5 & 3 & 0-2 & \textcolor{red}{.003} &.080 &.209 \\
    RP4-635E18.6 & 3 & 0-2 & \textcolor{red}{.003} &.094 &.183 \\
    RP4-635E18.8 & 3 & 0-2 & \textcolor{red}{<.001} &.092 &.092 \\
    RP4-739H11.3 & 3 & 0-2 & \textcolor{red}{<.001} &.097 &.149 \\
    RP4-758J18.7 & 3 & 0-1 & \textcolor{red}{<.001} &.262 &.146 \\
    RP5-1057J7.6 & 3 & 0-2 & \textcolor{red}{.013} &.185 &.185 \\
    RP5-1113E3.3 & 3 & 0-2 & \textcolor{red}{<.001} &.207 &.213 \\
    RP5-1125N11.1 & 3 & 0-2 & \textcolor{red}{.009} &.520 &.176 \\
    RP5-850O15.3 & 3 & 0-2 & \textcolor{red}{<.001} &.144 &.344 \\
    RP5-850O15.4 & 3 & 0-2 & \textcolor{red}{<.001} &.073 &.104 \\
    RP5-888M10.2 & 3 & 0-2 & \textcolor{red}{.004} &.139 &.184 \\
    RP5-892K4.1 & 3 & 0-2 & \textcolor{red}{<.001} &.183 &.200 \\
     \hline
\end{longtable}
\endgroup

\clearpage
\bibliographystyle{plainnat}
\bibliography{ref}

\end{document}